%% file: acl.tex
\pdfoutput=1

\documentclass[11pt]{article}

\usepackage{acl}

\usepackage{times}
\usepackage{latexsym}

\usepackage[T1]{fontenc}

\usepackage[utf8]{inputenc}

\usepackage{microtype}

\usepackage{inconsolata}

\input{math_commands.tex}

\usepackage{hyperref}
\usepackage{url}

\usepackage[utf8]{inputenc}
\usepackage{outlines}
\usepackage{graphicx}
\usepackage{mathtools}
\usepackage{amsmath}
\usepackage{amssymb}
\usepackage{cleveref}
\usepackage{mathtools}
\usepackage{makecell}
\usepackage{caption}
\usepackage{tabularx}
\usepackage{soul}

\newcommand{\bovae}{{Bo\hspace{-0.2ex}V-AE}}
\newcommand{\embtoemb}{\operatorname{Emb2Emb}}

\newcommand{\autoencoder}{\operatorname{\mathcal{A}}}

\newcommand{\transformer}{\operatorname{Transformer}}

\newcommand{\discretespace}{\mathcal{X}}
\newcommand{\embeddingspace}{\mathcal{Z}}

\newcommand{\encoder}{\mathrm{enc}}
\newcommand{\decoder}{\mathrm{dec}}
\newcommand{\loss}{\mathcal{L}}

\newcommand{\transformerplusplus}{\operatorname{Transformer\texttt{++}}}
\newcommand{\bov}[1]{\mathbb{#1}}
\newcommand{\hypothesisone}{\textbf{H1}}
\newcommand{\hypothesistwo}{\textbf{H2}}

\usepackage{accents}

\newcommand{\haussdorff}{\operatorname{H}}
\newcommand{\haussalign}{\operatorname{align}}
\renewcommand{\boldsymbol}{}
\newcommand{\bigO}{\mathcal{O}}

\newboolean{short}
\newboolean{showcomments}

\setboolean{short}{true}
\setboolean{showcomments}{false}

\newcommand{\short}[2]{\ifthenelse{\boolean{short}}{#1}{#2}}

\ifthenelse{\boolean{showcomments}}
{
\newcommand{\florian}[1]{\textcolor{cyan}{[Florian: #1]}}
\newcommand{\james}[1]{\textcolor{red}{[James: #1]}}
}{
\newcommand{\florian}[1]{}
\newcommand{\james}[1]{}
}

\title{Bag-of-Vectors Autoencoders for \\
Unsupervised Conditional Text Generation}

\author{Florian Mai \\
Idiap Research Institute / EPFL \\
Rue Marconi 19, 1920 Martigny \\
Switzerland \\
\texttt{florian.mai@idiap.ch}
\And
James Henderson \\
Idiap Research Institute \\
Rue Marconi 19, 1920 Martigny \\
Switzerland \\
\texttt{james.henderson@idiap.ch}
}

\begin{document}
\maketitle
\begin{abstract}
Text autoencoders are often used for unsupervised conditional text generation by applying mappings in the latent space to change attributes to the desired values.
Recently, \citet{mai2020plug} proposed $\embtoemb$, a method to \emph{learn} these mappings in the embedding space of an autoencoder. 
However, their method is restricted to autoencoders with a single-vector embedding, which limits how much information can be retained. 
We address this issue by extending their method to \emph{Bag-of-Vectors Autoencoders} (BoV-AEs), which encode the text into a variable-size bag of vectors that grows with the size of the text, as in attention-based models. 
This allows to encode and reconstruct much longer texts than standard autoencoders.
Analogous to conventional autoencoders, we propose regularization techniques that facilitate learning meaningful operations in the latent space. 
Finally, we adapt $\embtoemb$ for a training scheme that learns to map an input bag to an output bag, including a novel loss function and neural architecture.
Our empirical evaluations on unsupervised sentiment transfer \short{}{and sentence summarization }show that our method performs substantially better than a standard autoencoder.
\end{abstract}

\input{sections/introduction.tex}

\input{sections/background}

\input{sections/method.tex}

\ifthenelse{\boolean{short}}
{
\input{sections/experiments_short.tex}
}{
\input{sections/experiments_long.tex}
}

\input{sections/related.tex}

\input{sections/conclusion_short.tex}

\newpage

\bibliography{anthology,custom}
\bibliographystyle{acl_natbib}

\pagebreak

\appendix

\section*{Ethics Statement}
\label{sec:ethics}
\input{sections/ethics}

\input{sections/limitations.tex}

\section*{Reproducibility Statement}
\input{sections/reproducibility}

\input{sections/appendix}

\end{document}

%% file: math_commands.tex
\usepackage{amsmath,amsfonts,bm}

\def\eqref#1{equation~\ref{#1}}

\def\1{\bm{1}}

\DeclareMathAlphabet{\mathsfit}{\encodingdefault}{\sfdefault}{m}{sl}
\SetMathAlphabet{\mathsfit}{bold}{\encodingdefault}{\sfdefault}{bx}{n}

%% file: sections/introduction.tex
\section{Introduction}\label{sec:introduction}
In conditional text generation, we would like to produce an output text given an input text. Hence, parallel input-output pairs are required to train a good supervised machine learning model on this type of task. 
Large-scale pretraining~\citep{peters2018deep, devlin-etal-2019-bert, lewis2019bart} can alleviate the necessity for training examples to some extent, but even this requires a substantial number of annotations~\citep{yogatama2019learning}. This is an expensive process and can introduce unwanted artifacts itself, which are henceforth learned by the model~\citep{gururangan2018annotation}.
For these reasons, there is substantial interest in unsupervised solutions. \emph{Text autoencoders} (AEs) don't require labeled data for training, and are therefore a popular model for unsupervised approaches to many tasks, such as machine translation~\citep{artetxe2017unsupervised}, sentence compression~\citep{fevry2018unsupervised} and sentiment transfer~\citep{shen2017style}. The classical text AE~\citep{bowman2015generating} embeds the input text into a single fixed-size vector via the encoder, and then tries to reconstruct the input text from the single vector via the decoder. Single-vector embeddings are very useful, because they allow to perform conditional text generation through simple mappings in the embedding space, e.g.\ by adding a constant offset vector to change attributes such as sentiment~\citep{shen2019latent}. Recently, \citet{mai2020plug} proposed $\embtoemb$, a method that can \emph{learn} these mappings directly in the embedding space of any pretrained single-vector AE.
This is a powerful framework, because the AE can then be pretrained on unlimited amounts of unlabeled data before applying it to any downstream application. 
This concept, \emph{transfer learning}, is arguably one of the most important drivers of progress in machine learning in the recent decade: 
These so-called \emph{Foundation Models}~\citep{bommasani2021opportunities} have revolutionized natural language understanding (e.g, \emph{BERT}~\citep{devlin-etal-2019-bert}) and computer vision (e.g, \emph{DALL-E}~\citep{ramesh2021zero}), among others. 
Since $\embtoemb$ was designed to work with any pretrained AE, it was an important step towards their \emph{scalability}.

\begin{figure*}
    \centering
   \begin{minipage}{0.45\textwidth}
        \centering
            \includegraphics[width=\columnwidth]{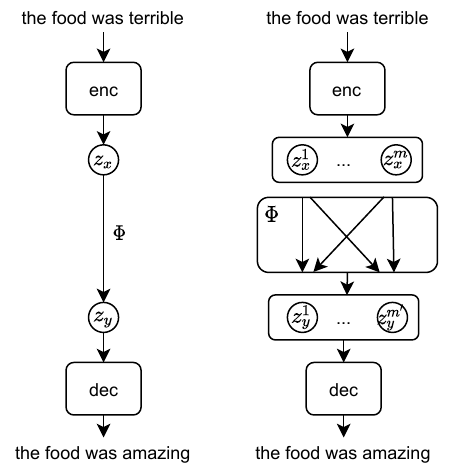}
        \caption{\emph{Left}: In the standard setup, the representation consists of a single vector, requiring a simple vector-to-vector mapping to do operations in the vector space. \emph{Right}: In BoV-AE, the representation consists of a variable-size bag of vectors, requiring a more complex mapping from one bag to another bag.}
        \label{fig:intro-autoencoders}
    \end{minipage}\hfill
    \begin{minipage}{0.45\textwidth}
        \centering
         \includegraphics[width=\columnwidth]{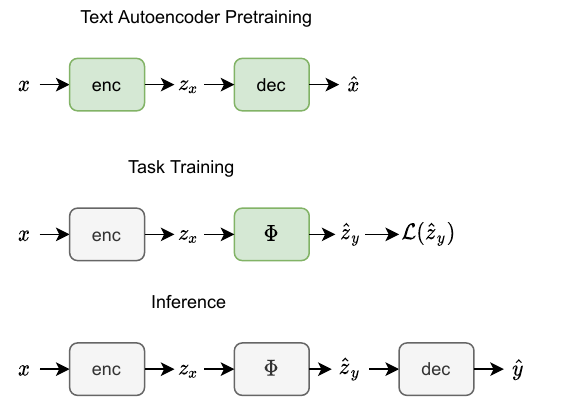}
        \caption{High-level view of the Emb2Emb framework.  \textit{Text Autoencoder Pretraining}:  An autoencoder is trained on an unlabeled corpus, i.e., the encoder $\encoder$ transforms an input text $x$ into a continuous embedding $\mathbf{z}_{x}$, which is in turn used by the decoder $\decoder$ to predict a reconstruction $\hat{x}$ of the input sentence.  \textit{Task Training}: The encoder is frozen (grey), and a mapping $\Phi$ is trained (green) on input embeddings $\mathbf{z}_{x}$ to output predictions $\mathbf{\hat{z}}_{y}$ such that it minimize some loss $\loss(\mathbf{\hat{z}}_{y})$. \textit{Inference}:  To obtain textual predictions $\hat{y}$, the encoder is composed with $\Phi$ and the decoder.}
        \label{fig:emb2emb-framework}
    \end{minipage}
    
\end{figure*}

However, as \citet{bommasani2021opportunities} point out, another crucial model property is \emph{expressivity}, the ability to represent the data distribution it is trained on.
In this regard, single-vector representations are fundamentally limited; they act as a bottleneck, causing the model to increasingly struggle to encode longer text~\citep{bahdanau2014neural}.
In this paper, we extend 
conditional text generation methods
from single-vector bottleneck AEs to \emph{Bag-of-Vector Autoencoders} (BoV-AEs), which encode text into a variable-size representation where the number of vectors grows with the length of the text.  This gives BoV-AEs the same kind of representations as attention-based models.
But this added expressivity comes with additional challenges: First, it can more easily overfit, leading to a non-smooth embedding space that is difficult to learn in. Secondly, as illustrated in Figure~\ref{fig:intro-autoencoders}, in the single-vector case, an operation $\Phi$ in the vector space consists of a simple vector-to-vector mapping, and a single-vector loss. But with BoV-AEs, $\Phi$ needs to map a bag of vectors onto another bag of vectors, for which the single-vector mapping and loss are not applicable. In this paper, we demonstrate how such a mapping can be learned in the context of the $\embtoemb$ framework by making the following novel contributions: \textbf{(i)} We propose a regularization scheme for BoV-AEs, \textbf{(ii)} a neural mapping architecture $\Phi$ for $\embtoemb$, and $\textbf{(iii)}$ a suitable training loss.

Empirically, we show on two unsupervised sentiment transfer datasets~\citep{shen2017style} of drastically different text lengths that BoV-AEs perform substantially better than standard AEs if the text is too long to be captured by one vector alone. Our ablation studies confirm that our technical contributions are crucial for this success.

In the following section, we review the $\embtoemb$ framework, before we introduce BoV-AE (Section~\ref{sec:bovae}) and its integration within $\embtoemb$ (Section~\ref{sec:emb2emb}).

%% file: sections/background.tex
\section{Background: Emb2Emb
}\label{sec:background}

\emph{Embedding-to-Embedding} ($\embtoemb$) \citep{mai2020plug} is a general framework for both supervised and an unsupervised conditional text generation. The core idea is to disentangle the specific task from the transition from the discrete text space to a continuous latent space (\emph{plug and play}), allowing for larger-scale pretraining with unlabeled data.

\ifthenelse{\boolean{short}}{}{\paragraph{Workflow}}
The workflow of the framework is depicted in Figure~\ref{fig:emb2emb-framework}.
First, a text AE $\autoencoder = \decoder \circ \encoder$ is trained to map an input sentence from the discrete text space $\discretespace$ to an embedding space $\embeddingspace$ via the encoder $\encoder : \discretespace \rightarrow \embeddingspace$, and back to $\discretespace$ via a decoder $\decoder : \embeddingspace \rightarrow \discretespace$, such that $\autoencoder(x) = x$, typically trained via negative log-likelihood, $\loss_{rec} = \operatorname{NLL}(\autoencoder(x), x)$. In contrast to other methods, $\autoencoder$ can in principle be any AE, opening the possibility for large-scale AE pretraining with unlabeled data.
Second, task-specific training is performed only in the embedding space $\embeddingspace$ of the AE. 
To this end, the encoder is frozen, and a new mapping layer $\Phi : \embeddingspace \rightarrow \embeddingspace$ is introduced, which is trained to transform the embedding of the input $\mathbf{z}_x$ into the embedding of the predicted output $\hat{\mathbf{z}}_y$.
The concrete loss $\loss(\hat{\mathbf{z}}_y)$ depends on the type of task.  In the supervised case, the true output is also encoded into space $\embeddingspace$, and the distance between the true embedding and the predicted embedding is minimized.
In the unsupervised case, the loss needs to be defined for the specific task at hand. 
For example, for sentiment transfer, where the goal is to transform a negative review into a positive review while retaining as much of the input as possible,
~\citet{mai2020plug} compose the loss as a combination of two loss terms\footnote{Their total loss includes an adversarial component that encourages the outputs of the mapping to stay on the latent space manifold. We leave adaptation of this component to the BoV scenario for future work.},
$\loss(\mathbf{\hat{z}}_y) = \loss_{sim}(\mathbf{z}_x, \mathbf{\hat{z}}_y) + \lambda_{sty} \loss_{sty}(\mathbf{\hat{z}}_y)$.
$\loss_{sty}$ encourages $\mathbf{\hat{z}}_y$ to be classified as a positive review according to a separately trained sentiment classifier. $\loss_{sim}$ encourages the output to be close to the input in embedding space, e.g. via euclidean distance.
$\lambda_{sty}$ is a hyperparameter that controls the importance of changing the sentiment of the predicted output.

\ifthenelse{\boolean{short}}{}{\paragraph{Choosing the embedding space}}
A main question in $\embtoemb$ is how to choose the embedding space $\embeddingspace$. \citet{mai2020plug} use a single continuous vector to encode all the information of the input, i.e. $\embeddingspace = \mathbb{R}^d$. This choice simplifies the mapping $\Phi$ to an MLP and the training loss to vector space distances, which is relatively easy to train. On the other hand, it limits the model in fundamental ways: The representation is \emph{fixed-sized}, i.e., the representation cannot grow in size. Sequence-to-sequence models with a fixed-size bottleneck struggle to encode long text sequences~\citep{bahdanau2014neural}, which is a key reason why attention-based models are now standard practice in sequence-to-sequence models.
Hence, it would be desirable to adapt $\embtoemb$ in such a way that $\embeddingspace$ contains \emph{variable-sized} embeddings instead.

%% file: sections/method.tex
\section{Bag-of-Vectors Autoencoder}\label{sec:bovae}

We propose \emph{Bag-of-Vectors Autoencoders} (BoV-AEs) 
which facilitate learning mappings in the embedding space.
Following the naming convention by \citet{henderson2020unstoppable}, we refer to a bag of vectors as a (multi)-set of vectors that (i) can grow arbitrarily large, and (ii) where the elements are not ordered (a basic property of sets).
A type of BoV representation that is used very commonly is found in Transformer~\citep{vaswani2017attention} encoder-decoder models, where there is one vector to represent each token of the input text, and the order of the vectors does not matter when the decoder accesses the output of the encoder.
In this work, we also rely on Transformer models as the backbone of our encoders and decoders. However, in principle, any encoder and decoder can be used, as long as the encoder produces a bag as output and the decoder takes a bag as input.
Formally, $\embeddingspace = (\mathbb{R}^d)^+$, so the encoder produces a bag-of-vectors $\bov{X} = \{\mathbf{z}_1, ..., \mathbf{z}_n\} := \encoder(x)$, where $n$ is the number of vectors in the induced input bag.

\subsection{Regularization}
The fact that we use a BoV-based AE presents a major challenge:
AEs have to be regularized to prevent them from learning a simple identity mapping where the input is merely copied to the output, which does not result in a meaningful embedding space. In fixed-size embeddings, this is for example achieved through under-completeness (choosing a latent dimension that is smaller than the input dimension) or through injection of noise, either at the input or in the embedding space. While there exists a lot of research on regularizing fixed-sized AEs, it is not clear how to achieve the same goal in a BoV-AE.
\ifthenelse{\boolean{short}}
{Here, regularizing the capacity of each vector is not enough. As long as each vector can store a (constant) positive amount of information, a bag of unlimited size can still store infinite information. However, it is not clear to what extent the size of the bag needs to be restricted. By default, a standard Transformer model produces as many vectors as there are input tokens, but this is likely too many, as it makes copying from the input to the output trivial. Hence, we want the encoder to output fewer vectors. In the following we explain how this is achieved in BoV-AEs.}
{Our main argument is that the nature of BoV models requires two types of regularization, (a) regularizing each vector in the bag and (b) regularizing the size of the bag.
The reasoning is as follows: As long as each vector can store a (constant) positive amount of information, a bag of unlimited size can store infinite information, ruling out that item (a) alone could be enough. On the other hand, even a single vector can in theory store any finite amount of information (and hence any text we observe in practice) if it is chosen large enough. Therefore, item (b) alone is not enough.

In practice, (a) can be achieved through under-completeness. Additionally, we leverage a denoising objective~\citep{vincent2010stacked}, which was recently shown to smoothen the embedding space of text AEs~\citep{shen2019latent}.
For (b), however, it is not straight-forward how to set the size of the bag. The standard Transformer model produces as many vectors as there are input tokens, but this is likely too many, as it makes copying from the input to the output trivial. Hence, we want the encoder to output fewer vectors. In the following we explain how this is achieved in BoV-AEs.}

Ideally, we want the model to decide for itself on a per-example basis which vectors it needs to retain for reconstruction.
To this end, we adopt \emph{L0Drop}, a differentiable approximation to L0 regularization, which was originally developed by \citet{zhang2020sparsifying} for the purpose of speeding up a model through sparsification.
The model computes scalar gates $g_i = g(\mathbf{z}_i) \in [0, 1]$ (which can be exactly zero or one) for each encoder output.
After the gates are computed, we multiply them with their corresponding vector. Vectors whose gates are near zero (i.e., smaller than some $\epsilon > 0$) are removed from the bag entirely\ifthenelse{\boolean{short}}
{.}
{:
\begin{equation*}
    X = \{ \boldsymbol{g}_i \odot \boldsymbol{z}_i : \boldsymbol{g}_i \geq \epsilon\}.
\end{equation*}
}
\ifthenelse{\boolean{short}}{An additional loss term, $\loss_{L_0}(\bov{X}) = \lambda_{L0} \sum_{i}^n g_i$ encourages the model to close as many gates as possible, where the hyperparameter $\lambda_{L0}$ controls the sparsity rate \emph{implicitly}. However, in initial experiments, we found $\lambda_{L0}$ difficult to tune, as it is very sensitive with respect to other hyperparameters.}
{An additional loss term encourages the model to close as many gates as possible:
\begin{equation*}
\loss_{L_0}(\bov{X}) = \sum\limits_{i}^n g_i.
\end{equation*}
\citet{zhang2020sparsifying} implicitly control the sparsity rate by introducing a hyperparameter $\lambda$ that acts as a weight for this additional loss term, in total: $\loss = \loss_{rec} + \lambda \loss_{L_0}$. However, in initial experiments, we found $\lambda$ difficult to tune, as it is very sensitive with respect to other hyperparameters.}
We instead employ a modified loss that seeks to \emph{explicitly} match a certain target ratio $r$ of open gates. Similar to the \emph{free-bits} objective that is used to prevent the posterior collapse problem in VAEs~\citep{kingma2016improved}, the objective becomes
\ifthenelse{\boolean{short}}{
\begin{equation}\label{eq:l0drop}\textstyle
\loss_{L_0}(\bov{X}) = \lambda_{L0} \max ( r, \frac{1}{n} \sum_{i}^n g_i ).
\end{equation}
}{
\begin{equation*}
\loss_{L_0}(\bov{X}) = \lambda_{L0} \max \left( r, \frac{1}{n} \sum\limits_{i}^n g_i \right).
\end{equation*}}
By setting $\lambda_{L0}$ to a large enough value (empirically, $\lambda_{L0} = 10$), we find that this objective reaches the target ratio $r$ reliably for different $r$ while at the same time reducing the reconstruction loss. This allows to compare different strengths of regularization while reducing the tuning effort substantially.

\section{Emb2Emb with BoV-AEs}\label{sec:emb2emb}

In the following we describe how to adapt the Emb2Emb model to BoV-AEs, i.e., how to generate an output bag $\hat{\bov{X}} = \{\hat{\mathbf{z}}_1, \dots, \hat{\mathbf{z}}_n\}$ given an input bag $\bov{X}$ through the mapping $\Phi(\bov{X})$, and how to choose the loss function $\loss(\hat{\bov{X}},\bov{X})$. For example, in the case of style transfer, we want $\hat{\bov{X}}$ to be similar to $\bov{X}$. 

\subsection{Mapping $\Phi$}
In contrast to \citet{mai2020plug}, who use a single-vector embedding and hence $\Phi$ can be as simple as an MLP, in our work, $\Phi$ must be capable of producing a bag of vectors. The straight-forward choice for $\Phi$ is a Transformer decoder that uses cross-attention on the input BoV, and generates vectors autoregressively
one at a time, formally $\hat{\mathbf{z}} = \transformer(\mathbf{z}_s, \hat{\mathbf{z}}_{1}, \dots, \hat{\mathbf{z}}_{t-1}, \bov{X}), t \geq 1$, where $\mathbf{z}_s$ is the embedding of some starting symbol.  Since the resulting sequence of vectors is still interpreted as a bag by the decoder and loss function, the ordering is irrelevant, but generating vectors autoregressively facilitates modelling the correlations between vectors.

Depending on the difficulty of the task, a generic Transformer decoder may be sufficient to learn the mapping, but for more difficult mappings and for larger bags (i.e.\ longer texts) appropriate inductive biases are needed.
Based on the assumption that the output should be close to the input in embedding space, \citet{mai2020plug} propose \emph{OffsetNet} for the single vector case, which computes an offset vector to be added to the input. With a similar motivation, we propose a variant of pointer-generator networks~\citep{see2017get}, which allows the model to choose between copying an input vector and generating a new one. Instead of just copying, however, our model ($\transformerplusplus$) allows to compute an offset vector to be added to the copied vector, analogous to \cite{mai2020plug}. Formally, at each timestep $t$,
\begin{equation}
    \hat{\mathbf{z}}_t = (1 - p_{gen}) (\mathbf{z}_{copy} + \mathbf{z}_{\text{offset}}) + p_{gen} \mathbf{z}'_t,
\end{equation}
where $\mathbf{z}'_t = \transformer(\mathbf{z}_s,..., \hat{\mathbf{z}}_{t-1}, \mathbb{X})$.
Intuitively, by controlling $p_{gen} \in (0,1)$, the model makes the (soft) decision to either copy a vector from the input and add an offset, or to generate a completely new vector.
Here, $p_{gen}$ is a function of $\mathbf{z}'_t$ and the starting symbol which we treat as a context vector, $p_{gen} = \sigma(\mathbf{W}[\mathbf{z}_s; \mathbf{z}'_{t}])$. Similarly, $\mathbf{z} _{\text{offset}}$ is a one-layer MLP with $[\mathbf{z}'_{t}; \mathbf{z}_{copy}]$ as input. $\mathbf{z}_{copy}$ is determined through an attention function:
\begin{align}
    \mathbf{z}_{copy} = \sum\limits_{i = 1}^{|\bov{X}|} \mathbf{\alpha}_i \mathbf{z}_i, \quad
    \mathbf{K} = \mathbf{W}_{cpy} \mathbf{X}, \\
    \mathbf{\alpha}_i = \operatorname{softmax}(\mathbf{z}_s^T \mathbf{K})_i, \quad (\mathbf{X})_i := \mathbf{z}_i
\end{align}
where $\mathbf{W}_{cpy}$ is a learnable weight matrix. We refer to this model as $\transformerplusplus$.

\subsection{Generating Variable Sized Bags}\label{sec:method-variable-size-bag}

The output bag is generated in an autoregressive manner. In the unsupervised case, it is not always clear how many vectors the bag should contain. However, due to the unsupervised nature, all information needed for computing the (task-dependent) training loss $\mathcal{L}(\bov{\hat{X}},\bov{X})$ are also available at inference time. In this case, we can first generate some fixed maximum number $N$ of vectors autoregressively, and then determine the optimal bag by computing the minimal (inference-time) loss value, $\bov{X}^* = \min\limits_{l = 1,...,N} \mathcal{L}(\hat{\bov{X}}_{1:l}, \bov{X})$. This can be valuable for tasks where we do not have a good prior on the size of the target bag.
During training, we minimize the loss locally at every step\ifthenelse{\boolean{short}}
{. But we don't necessarily care about the loss at very small or big bags, so we might want to weight the steps as $\mathcal{L}^{\text{total}}(\hat{\bov{X}},\bov{X}) = \sum_{l = 1}^N \mathbf{w}_l \mathcal{L}(\hat{\bov{X}}_{1:l}, \bov{X})$.}
{:%

\begin{equation}
    \mathcal{L}^{\text{total}}(\bov{X},\bov{Y}) = \sum\limits_{l = 1}^N (\bov{X}_{1:l}, \bov{Y}).
\end{equation}

But we don't necessarily care about the loss at very small or big bags, so we might want to weight the steps
\begin{equation}
    \mathcal{L}^{\text{total}}(\bov{X},\bov{Y}) = \sum\limits_{l = 1}^N \mathbf{w}_l \mathcal{L}(\bov{X}_{1:l}, \bov{Y}).
\end{equation}}
Here, $\mathbf{w} \in \mathbb{R}_{+}^{N}$ could be any weighting, but it is more beneficial for training to only backpropagate from bag sizes that we expect to be close to the optimal output bag size. For instance, in style transfer, the output typically has about the same length as the input. Hence, for an input size of length $n$, a useful weighting could be
\begin{equation}\label{eq:windowsize}
    \mathbf{w}_l = \begin{cases}
1 & n - k \leq l \leq n + k \\
0 & \text{otherwise}
\end{cases},
\end{equation}
where $k$ is the size of a window around the input bag size.

\subsection{Aligning Two Bags of Vectors}\label{sec:hausdorff-sim}
As described in Section~\ref{sec:background}, unsupervised sentiment transfer involves two loss terms, $\loss_{sty}$ and $\loss_{sim}$. In order to adapt $\loss_{sty}$ from the single vector case to the BoV case, we can simply switch from an MLP classifier to a Transformer-based classifier. For $\loss_{sim}$, however, we need to switch to a loss function that is defined on sets. While there are well-known losses for the single-vector case,
in NLP set-level loss functions are not well-studied. 
\ifthenelse{\boolean{short}}{}
{Here, we adopt the Hausdorff distance, which is more commonly used in vision applications (e.g.~\cite{fan2017point}), and propose a novel variant.}

\ifthenelse{\boolean{short}}{Here, we propose a novel variant of the \emph{Hausdorff} distance. This distance is commonly used in vision applications: as a performance evaluation metric in e.g. medical image segmentation~\citep{taha2015metrics, aydin2020usage}, or in vision systems as a way to compare images~\citep{huttenlocher1993comparing, takacs1998comparing, lin2003spatially, lu2001approach}.
More recently, variants (different from ours) of the Hausdorff distance have also been used as loss functions to train neural networks~\citep{fan2017point,ribera2019locating, zhao2021phdlearninglearning}. In NLP, its use  is very rare~\citep{nutanong2016scalable, chen2019learning, kuo2020compositional}. To the best of our knowledge, our paper is the first to present a novel, fully differentiable variant of the Hausdorff distance as a loss for language learning.}{Here, we propose a novel variant of Hausdorff distance, which is more commonly used in vision applications (e.g.~\cite{fan2017point}), and propose a novel variant.}

\ifthenelse{\boolean{short}}{}{\paragraph{Hausdorff Distance}}
The Hausdorff distance is a method for aligning two sets. Given two sets $\bov{X}$ and $\hat{\bov{X}}$, their Hausdorff distance $\haussdorff$ is defined as
\ifthenelse{\boolean{short}}
{
\begin{align}
\haussdorff(\mathbb{X}, \hat{\bov{X}}) = \frac{1}{2} \haussalign(\mathbb{X},\hat{\bov{X}}) + \frac{1}{2}     \haussalign(\hat{\bov{X}},\mathbb{X}) \\
\haussalign(\mathbb{X}, \hat{\bov{X}}) = \max\limits_{x \in \mathbb{X}} \min\limits_{y \in \hat{\bov{X}}} \operatorname{d}(x,y) \label{eq:both-sided-hausdorff}
\end{align}
}
{
\begin{align}
    \haussalign(X, Y) &= \max\limits_{x \in X} \min\limits_{y \in X} \operatorname{d}(x,y) \\
    \haussdorff(X, Y) & = \frac{1}{2} \haussalign(X,Y) + \frac{1}{2} \haussalign(Y,X). \label{eq:both-sided-hausdorff}
\end{align}
}
Intuitively, two sets are close if each point in either set has a counterpart in the other set that is close to it according to some distance metric $d$. We choose $d$ to be the euclidean distance, but in principle any differentiable distance metric could be used (e.g. cosine distance).
\ifthenelse{\boolean{short}}
{However, the vanilla Hausdorff distance is very prone to outliers, and therefore often reduced to the \emph{average Hausdorff distance}~\citep{dubuisson1994modified}, where
\vspace{-1ex}
\begin{align}
\haussalign(\mathbb{X},\hat{\bov{X}}) & = \frac{1}{|\mathbb{X}|}\sum\limits_{x \in \mathbb{X}} \min\limits_{y \in \hat{\bov{X}}} d(x,y). \label{eq:avg-hausdorff}
\end{align}
The average Hausdorff function is step-wise smooth and differentiable. Empirically, however, we find step-wise smoothness to be insufficient for the best training outcome. Therefore, we propose a fully differentiable version of the Hausdorff distance by replacing the $\min$ operation with $\operatorname{softmin}$ by modelling $\haussalign(\mathbb{X}, \hat{\bov{X}}) =$
\vspace{-1ex}
\begin{equation}
 \frac{1}{|\mathbb{X}|}\sum\limits_{x \in \mathbb{X}} \sum\limits_{y \in \hat{\bov{X}}} \left( \frac{e^{(-d(x,y))}}{\sum\limits_{y' \in \hat{\bov{X}}} e^{(-d(x,y'))}} \cdot d(x,y) \right). \label{eq:diff-hausdorff}
\end{equation}
This variant is reminiscent of the attention mechanism~\cite{bahdanau2014neural} in the sense that a weighted average is computed, which has been very successful at smoothly approximating discrete decisions, e.g., read and write operations in the Differentiable Neural Computer~\citep{graves2016hybrid} among many others.}
{\paragraph{Average Hausdorff Distance}
The vanilla Hausdorff distance is very prone to outliers, and therefore often reduced to the average Hausdorff distance~\citep{dubuisson1994modified}, where
\begin{align}
\haussalign(\mathbb{X},\mathbb{Y}) & = \\ \frac{1}{|\mathbb{X}|}\sum\limits_{x \in \mathbb{X}} \min\limits_{y \in \mathbb{Y}} d(x,y). \label{eq:avg-hausdorff}
\end{align}
The average Hausdorff function is step-wise smooth and differentiable.

\paragraph{Differentiable Hausdorff}
Empirically, we find step-wise smoothness to be insufficient for the best training outcome. Therefore, we propose a fully differentiable version of the Hausdorff distance by replacing the $\min$ operation with $\operatorname{softmin}$ like follows:
\begin{equation}
\haussalign(\mathbb{X}, \mathbb{Y}) = \frac{1}{|\mathbb{X}|}\sum\limits_{x \in \mathbb{X}} \sum\limits_{y \in \mathbb{Y}} \left( \frac{e^{(-d(x,y))}}{\sum\limits_{y' \in \mathbb{Y}} e^{(-d(x,y'))}} \cdot d(x,y) \right). \label{eq:diff-hausdorff}
\end{equation}
This variant is reminiscent of the attention mechanism~\cite{bahdanau2014neural} in the sense that a weighted average is computed, which has been very successful at smoothly approximating discrete decisions, e.g., read and write operations in the Differentiable Neural Computer~\citep{graves2016hybrid} among many others.
}

%% file: sections/experiments_short.tex
\section{Experiments}
Our experiments are designed to test the following two hypotheses.
\hypothesisone: If the input text is too long to be encoded into a fixed-size single vector representation, \bovae-based $\embtoemb$ provides a substantial advantage over the fixed-sized model.
\hypothesistwo: Our technical contributions, namely L0Drop regularization, the training loss, and the mapping architecture, are necessary for \bovae's success.

We evaluate our model on two unsupervised conditional text generation tasks: In Section~\ref{sec:exp-yelp-reviews}, we show that \hypothesisone
~holds even when the single-vector dimensionality is large ($d {=} 512$). To this end, we create a new sentiment transfer dataset, Yelp-Reviews, whose inputs are relatively long.
However, training on this dataset is computationally very demanding\footnote{Pretraining a model of this size until convergence took more than a month on a single 24GB GPU.}.
Therefore, we turn to a short-text style transfer dataset to test hypothesis \hypothesistwo~(Section~\ref{sec:exp-sentence-yelp}).

Additionally, we conducted experiments on abstractive sentence summarization~\citep{rush2015neural}. These provide evidence of the generality of our method, as well as the utility of the mapping architecture's copy mechanism. Due to space constraints, these are included in Appendix~\ref{app:sentence-summarization}.

For each of the experiments in this section, we provide full experimental details in Appendix~\ref{app:exp-details}.

\ifthenelse{\boolean{short}}{\textbf{Evaluation metrics:}}{\paragraph{Evaluation metrics}}
In sentiment transfer, the goal is to rewrite a negative review as a positive review while keeping as much of the content as possible. Hence, two metrics are important, sentiment transfer ability and content retention. Following common practice~\citep{hu2017toward, shen2017style, subramanian2018multiple}, we measure the former with a separately trained style classifier based on DistilBERT~\citep{sanh2019distilbert}, and content retention in terms of self-BLEU~\citep{papineni2002bleu} between the input and the predicted output.
To allow comparison via a single score, we aggregate content retention and transfer accuracy~\citep{xu2018unpaired, krishna2020reformulatingunsupervised}, per sentence~\citep{krishna2020reformulatingunsupervised},
\ifthenelse{\boolean{short}}
{and compute a single $score = \frac{1}{M} \sum_{i = 1}^{M} \operatorname{ACC}(\hat{y}) \cdot \operatorname{BLEU}(\hat{y}, x)$}
{:
\begin{equation}\label{eq:style-transfer-score}
    score = \frac{1}{M} \sum\limits_{i = 1}^{M} \operatorname{ACC}(\hat{y}) \cdot \operatorname{BLEU}(\hat{y}, x)
\end{equation}}
where $x$ is the input sentence, $\hat{y}$ is the predicted sentence, and $M$ is the number of data points. 
For readability, we multiply all metrics by 100 before reporting.

\short{}{\input{sections/sentence-summarization-eval}}

\ifthenelse{\boolean{short}}{\textbf{Autoencoder Pretraining:}}{\paragraph{Autoencoder Pretraining}}
Since $\embtoemb$ is plug and play, the autoencoder pretraining can be decoupled from the downstream task, enabling large-scale pretraining on a general purpose corpus. While this would certainly be necessary to reach the best results possible, such an endeavor is very resource-intensive, making it impractical to conduct the kind of controlled experiments needed to support our hypotheses.
Moreover, existing pretrained autoencoders such as BART~\citep{lewis2019bart} cannot be used off-the-shelf because they weren't trained to have a smooth embedding space, for example using L0Drop.
In Appendix~\ref{app:yelp-sentences-pretraining}, we study the effect of adding an L0Drop layer inside BART and finetuning it for a few steps on the target task data. Although this works to some extent, this L0Drop layer can be expected to remove information which would be kept if it were trained in full large-scale pretraining, which we don't have the resources to do.

Therefore, we instead pretrain all autoencoders from scratch directly on the data of the target task. Models named $\textbf{L0-r}$ denote L0Drop-based BoV-AE models that only differ in the target ratio $r$ used in training. 
As a control, we always compare to a single vector \textbf{fixed}-size AE, which is obtained by averaging the vectors at the last layer of the encoder. 

\subsection{Yelp-Reviews}\label{sec:exp-yelp-reviews}

\begin{figure*}[t]
    \vspace{-2ex}
    \centering
    \begin{minipage}{0.45\textwidth}
        \centering
         \includegraphics[width=\columnwidth]{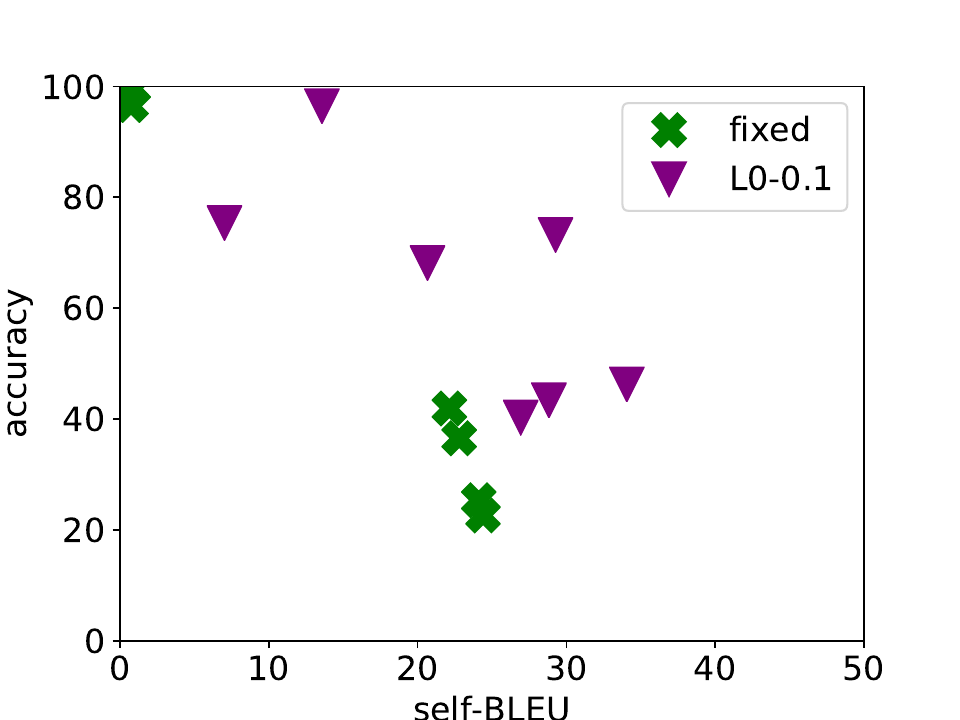}
        \caption{Style transfer on Yelp-Reviews.}
    \label{fig:exp-yelp-reviews-downstream}
    \end{minipage}\hfill
    \begin{minipage}{0.45\textwidth}
    \includegraphics[width=\textwidth]{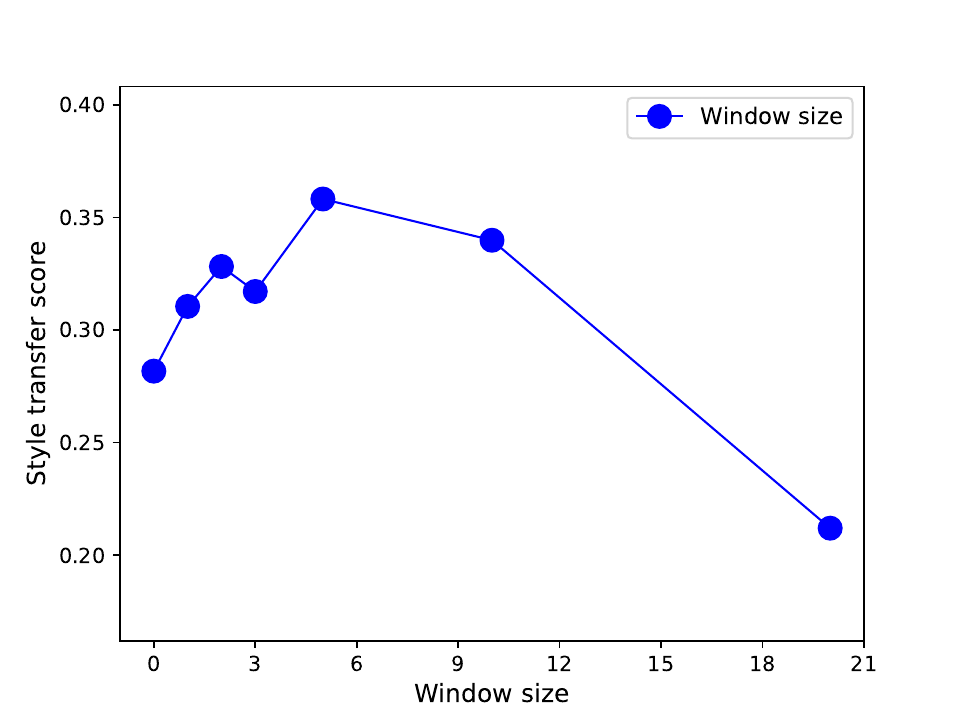}
    \caption{Style transfer score depending on the window size.}
    \label{fig:exp-window-size}

    \end{minipage}\hfill
    \begin{minipage}{0.45\textwidth}

        \centering
        \includegraphics[width=\columnwidth]{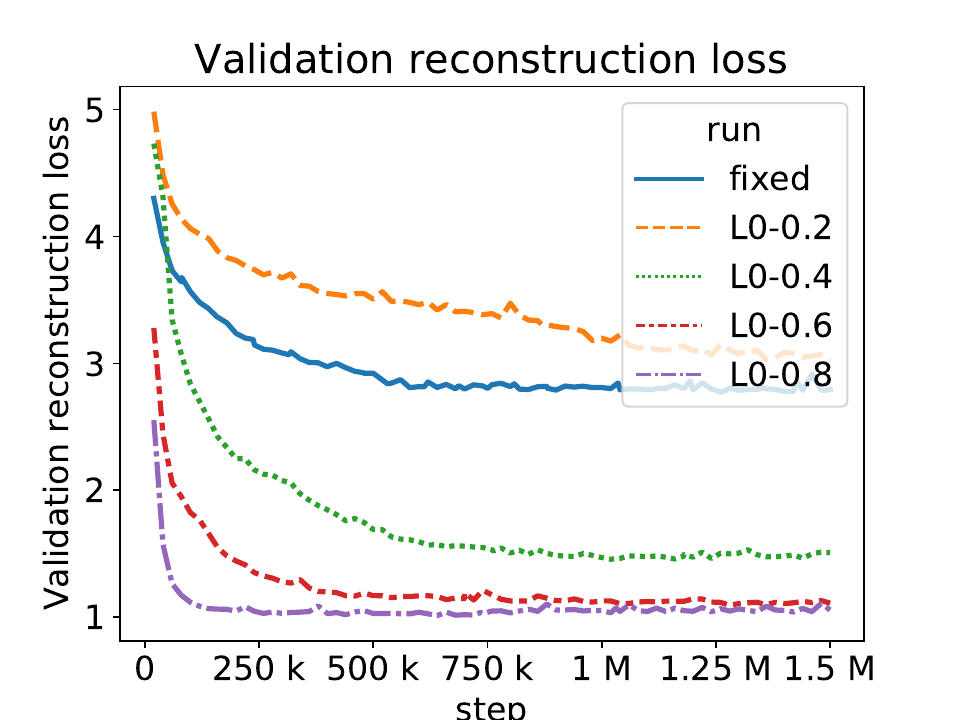}
        \caption{Reconstruction loss on the validation set for different AEs. \textbf{fixed}: A single vector obtained by averaging the encoder output vectors. \textbf{L0-r}: BoV-AEs with L0Drop target ratio $r$.}
        
    \label{fig:val_rec_loss}
    \end{minipage}\hfill
    \begin{minipage}{0.45\textwidth}
        \centering
     \includegraphics[width=\columnwidth]{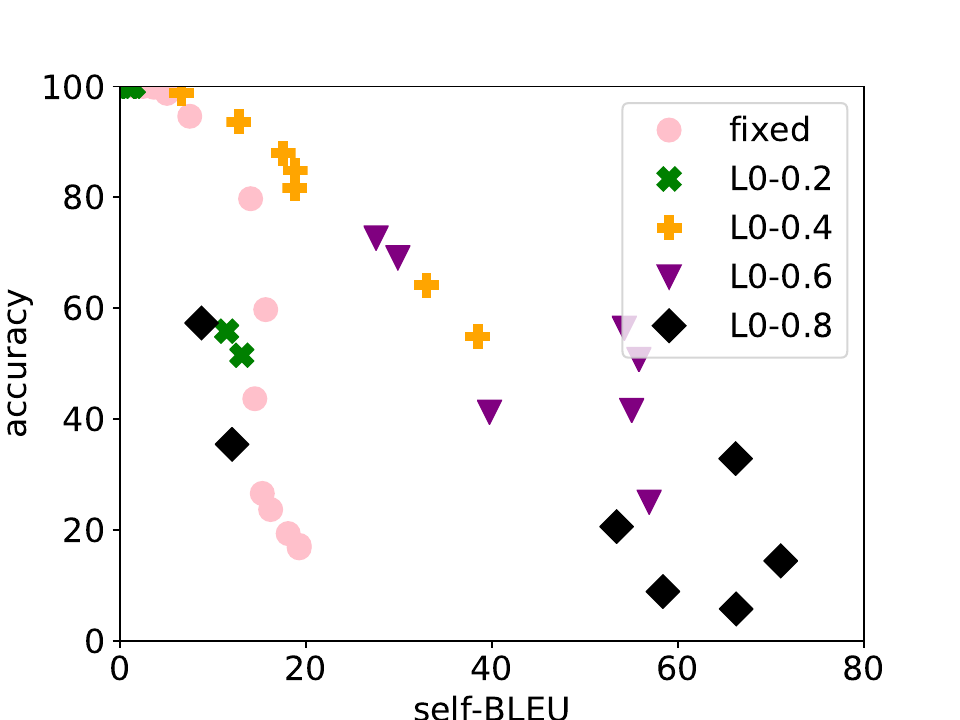}
        \caption{Style transfer performance on Yelp-Sentences of BoV models compared to a fixed-size AE for varying $\lambda_{sty}$.
        Further to the top (style transfer) and right (content retention) is better.}
        \label{fig:sentence-yelp-l0drop}
    \end{minipage}
        \vspace{-2ex}
\end{figure*}

Our hypothesis is that AEs with a single vector bottleneck are unable to reliably compress the text when it is too long.
Here, we test if this holds true even for a large single-vector model with $d {=} 512$. 
To this end, we create the dataset \emph{Yelp-Reviews}, which consists of strongly positive and strongly negative English restaurant reviews on Yelp (see Appendix~\ref{sec:app-yelp-reviews-dataset} for a detailed description).
This dataset is very similar to Yelp-Sentences introduced by \citet{shen2017style}. However, while Yelp-Sentences consists of single sentences of about 10 words on average, Yelp-Reviews consists of entire reviews of 52 words on average.
\ifthenelse{\boolean{short}}
{For style transfer, we train a $\transformerplusplus$ mapping using
the loss described in Section~\ref{sec:background}. To obtain results at varying transfer levels, we train multiple times with varying $\lambda_{sty}$, resulting in multiple points for each model in Figure~\ref{fig:exp-yelp-reviews-downstream} and \ref{fig:sentence-yelp-l0drop}.}
{To test the performance of BoV models on style transfer, we train a mapping using the loss term $\loss(\mathbf{\hat{z}}_y) = \loss_{sim}(\mathbf{z}_{x}, \mathbf{\hat{z}}_y) + \lambda_{sty} \loss_{sty}(\mathbf{\hat{z}}_y)$ as described in Section~\ref{sec:background}. To obtain results at varying transfer levels, we train multiple times with varying $\lambda_{sty}$.}

\ifthenelse{\boolean{short}}{\textbf{Results:}}{\paragraph{Results}}
The results (full graph shown in Figure~\ref{fig:yelp-reviews-valrec} in the Appendix) indicate that even large single vector models ($d {=} 512$) are unable to compress the text well;  the NLL loss on the validation set of the fixed-size model is ${{\approx}} 3.9$. $\textbf{L0-0.05}$ is only slightly better than the fixed-size model, whereas $\textbf{L0-0.1}$ already reaches a substantially lower reconstruction loss (${\approx} 2.1$). 
We evaluated the downstream sentiment transfer performance of $\transformerplusplus$ with $\textbf{L0-0.1}$\footnote{We restrict our analysis to L0-0.1 because this dataset have is computationally demanding.} and the fixed-size model, respectively.
Figure~\ref{fig:exp-yelp-reviews-downstream} shows a scatter plot of the results, where results that are further to the top-right corner are better. We see that at a comparable transfer level, the BoV is substantially better at retaining the input content.
This supports hypothesis $\hypothesisone$ that variable-size BoV models are particularly beneficial in cases where the text length is too long to be encoded in a single-vector. 

\subsection{Yelp-Sentences}\label{sec:exp-sentence-yelp}
In order to answer research question \hypothesistwo, we perform a large set of controlled experiments over our model's components. 
Due to the high computational demand, we turn to the popular Yelp-Sentences sentiment transfer dataset by \citet{shen2017style}. Texts in this dataset are ${\approx}$ 10 words on average. As these sentences are much easier to reconstruct, we set the embedding size to $d {=} 32$
so that the condition for hypothesis \hypothesisone\ is still valid.
Here, we again train BoV-AEs for a variety of target rates ($r = 0.2, 0.4, 0.6, 0.8$) and then evaluate their reconstruction and style transfer ability in the same fashion as for Yelp-Reviews. Finally, we investigate the impact of the differentiable Hausdorff loss and the window size. 
For completeness, we provide an analysis of the computational complexity of BoV-AE in Appendix~\ref{app:yelp-sentences-time}.

\subsubsection{Reconstruction Ability}
Figure~\ref{fig:val_rec_loss} shows the reconstruction loss on the validation set for the fixed-size model compared to BoV models.
The fixed-size AE does not reach satisfactory reconstruction ability, converging at an NLL loss value of about 3. In contrast, BoV models are able to outperform the fixed-size model considerably. As expected, higher target ratios lead to better reconstruction, because the model can use more vectors to store the information. Models with a higher target ratio also reach their optimal loss value more quickly. While $\textbf{L0-0.6}$ approaches the best reconstruction value (${\approx} 1.0$) eventually, the model needs more than 1 million training steps to reach it. In contrast, $\textbf{L0-0.8}$ needs less than 100k steps to converge, which could indicate that $\textbf{L0-0.8}$ learns to copy rather then compress the input, resulting in a bad latent space.
$\textbf{L0-0.4}$ yields to a higher loss, but is still drastically better than the fixed size model. $\textbf{L0-0.2}$ is not enough to outperform the fixed-size model. Overall, these results show we have the right settings for evaluating \hypothesisone\ and \hypothesistwo, as 10 words is too long to be encoded well into a single vector of $d {=} 32$, whereas a BoV-AE with a high enough target ratio $r$ can fit it well.

\subsubsection{Style Transfer Ability}
Results are shown in Figure~\ref{fig:sentence-yelp-l0drop}. Up to $r {=} 0.6$, they correspond well to the reconstruction ability, in that BoV models with higher target ratios yield higher self-BLEU scores at comparable transfer abilities, outperforming the fixed-size model (\hypothesisone). 
However, at $r {=} 0.8$, the performance suddenly deteriorates at medium to high transfer levels. This supports the hypothesis that $\textbf{L0-0.8}$ lacks smoothness in the embedding space due to insufficient regularization, which in turn complicates downstream training. This is the first piece of evidence that L0Drop is necessary for the success of our model (\hypothesistwo).

\subsubsection{Ablation on Differentiable Hausdorff}
In Section~\ref{sec:hausdorff-sim}, we argue that the min operation should be replaced by softmin in order to facilitate backpropagation. Here, we test if the differentiable version is really necessary, that is, we compare Eq.~\ref{eq:avg-hausdorff} to Eq.~\ref{eq:diff-hausdorff}. Like above, we train the two variants with different $\lambda_{sty}$, and then select the best style transfer score on the validation set. 
The difference is substantial: Average Hausdorff reaches 14.6, whereas differentiable Hausdorff reaches 24.2. We hypothesize that this discrepancy is due to the difficult nature of the style transfer problem, which requires carefully balancing the two objectives, content retention (via Hausdorff) and style transfer (via the classifier). This is easier when the objective functions are smooth, which is the advantage of differentiable Hausdorff.

\subsubsection{Ablation on Window Size}
\input{sections/window-size}

\iffalse
\ifthenelse{\boolean{short}}{\textbf{Using Pretrained Autoencoders:}}{\paragraph{Using Pretrained Autoencoders}}
A unique advantage of the $\embtoemb$ framework is its compatibility with pretrained autoencoders. In order to show that our BoV extension also benefits from pretraining, we build on top of BART~\cite{lewis2019bart}, which uses similar resources as BERT, but is trained via a denoising autoencoder objective. However, we find that plain BART does not work well with $\embtoemb$. This is expected, as BART does not regularize the size of the latent space. Therefore, we add an L0Drop layer after the BART encoder and finetune it for few steps on Yelp-Sentences. Our experiments confirm that L0Drop is strictly necessary to make the model work, and that the quality of outputs is better compared to no pretraining. The full analysis can be found in Appendix~\ref{app:yelp-sentences-pretraining}.
\fi

\short{}{\input{sections/sentence-summarization}}

\input{sections/qualitative-analysis}

%% file: sections/window-size.tex
The window size $k$ determines which bag sizes around the input bag size we backpropagate from (cmp. Section~\ref{sec:method-variable-size-bag}). Here, we investigates its influence on the model's performance. Since the $\lambda_{sty}$ hyperparameter is very sensitive to other model hyperparameters, we train with varying $\lambda_{sty}$ for each fixed window size and report the best style transfer score for each window size. 
In Figure~\ref{fig:exp-window-size}, we plot the style transfer score as a function of the window size.
Our results indicate that increasing the window size from zero (score 28.2) is beneficial up to some point ($k {=} 5$, score 35.8), whereas increasing by too much ($k {=} 20$, score 21.2) is detrimental to model performance even compared to a size of zero. We hypothesize that backpropagating bags that are either very small or very large is detrimental because it forces the model to adjust its parameters to optimize unrealistic bags, taking away capacity for fitting realistic bags.

%% file: sections/sentence-summarization.tex
\subsection{Experimental Setup}
In sentence summarization~\citep{rush2015neural}, the goal is to capture the essence of a sentence in fewer words. 
We evaluate on the Gigaword corpus~\citep{graff2003english} similar to \citet{rush2015neural}. This corpus consists of more than 8.5 million training samples, but we use a random subset of 500k to limit the computational cost.
Inputs are on average 27 words long, which is medium length compared to the other two datasets in this study. We use moderately sized vectors of $d {=} 128$ and again train different BoV-AEs with target ratios $r = 0.2, 0.4, 0.6, 0.8$.
When applying the model to the sentence summarization downstream task, we train using the loss term $\loss(\mathbf{\hat{z}}_y) = \loss_{sim}(\mathbf{z}_{x}, \mathbf{\hat{z}}_y) + \lambda_{len} \loss_{len}(\mathbf{\hat{z}}_y)$. 
This loss term is conceptually similar to the loss term used for style transfer, except that $\loss_{len}$ denotes the prediction of a model trained to predict the length of the input text from the text's latent representation (the shorter the better). We train with varying values of $\lambda_{len} = 0.1, 0.2, 0.5, 1, 2, 5, 10$ and select the best model (ROUGE-L) on the development set.
Intuitively, this model learns to retain as much from the input as possible while minimizing the output length. Note that this model of summarization could certainly be improved further, e.g.\ by accounting for relevancy and informativeness of the output~\citep{peyrard2018simple}. However, our goal is not to create the best task-specific model possible, so these considerations are out of scope for this paper.

The input texts in this task are relatively long. Due to the higher number of vectors in a BoV, it may be difficult to learn the mapping, especially for large target ratios $r$.
We experiment with $\transformerplusplus$ to observe to what extent this can facilitate learning.

As is standard practice in summarization, we evaluate performance on this task with ROUGE-L~\citep{lin2004rouge}.
Note, however, that ROUGE scores can be misleading, because even texts that are as long or even longer than the input text can yield relatively high scores even though they are clearly not summaries. For this reason, we also report the average length of outputs produced by the models as reference.

\subsection{Results}
\begin{figure*}
    \centering    
    \includegraphics[width=\columnwidth]{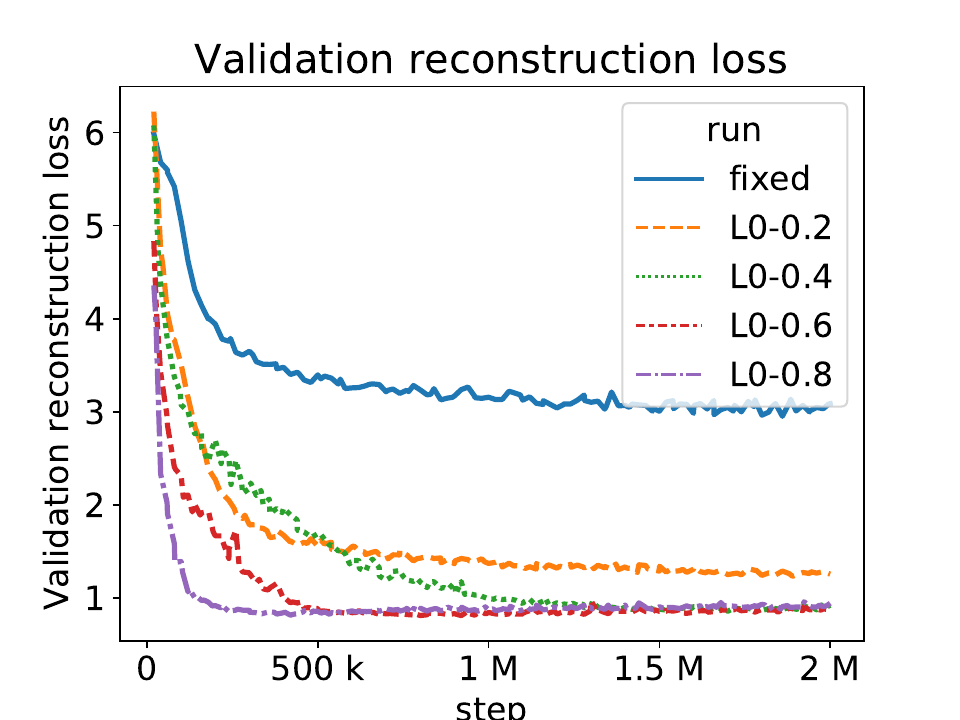}
    \caption{Reconstruction loss on the validation set of Gigaword for different autoencoders. \textbf{fixed}: The bag consists of a single vector obtained by averaging the embeddings at the last layer of the Transformer encoder. \textbf{L0-r}: BoV-AE with L0Drop target ratio $r$.}
    \label{fig:abssum-valrec}
\end{figure*}

Figure~\ref{fig:abssum-valrec} shows the development of the reconstruction loss on the validation set over the course of 2 million training steps.
Despite the moderately large vector dimensionality, the single-vector bottleneck model achieves only considerably lower reconstruction performance than the BoV models. Again, larger target rates $r$ lead to faster convergence, and all BoV models converge to approximately the same validation loss value (0.9). The only exception is \textbf{L0-0.2}, which converges to a higher loss value (1.25), but is still vastly stronger than the fixed size model (3.01).

\begin{table}[t]
\centering
    \captionof{table}{Results on Gigaword sentence summarization. Scores represent ROUGE-L with average output words in parentheses. T and T\texttt{++} denote $\transformer$ and $\transformerplusplus$, respectively.}
    \label{tab:res-sentsum}
    \begin{tabular}{c||c|c}
    Model & $\operatorname{T}$ & $\operatorname{T\texttt{++}}$ \\
    \hline
         fixed & 13.1 (\textit{18.3})  & 13.2 (\textit{17.6}) \\
         \hline
         L0-0.2 & 19.8 (\textit{23.2}) & 18.3 (\textit{10.7}) \\
         L0-0.4 & 8.0 (\textit{18.7}) & 16.4 (\textit{12.5})  \\
         L0-0.6 & 6.6 (\textit{83.5}) &  14.7 (\textit{51.1}) \\
         L0-0.8 & 9.3 (\textit{5.1}) &  13.2 (\textit{48.6}) \\
    \end{tabular}

\end{table}

However, as shown in Table~\ref{tab:res-sentsum},
\textbf{L0-0.2} performs the best on the downstream task, outperforming the single-vector model by more than 5 ROUGE-L points while simultaneously requiring much fewer output words. 
BoV models with higher target ratios than $r {=} 0.2$ perform worse. Moreover, the $\transformerplusplus$ architecture tends to improve results, particularly with target rates $r > 0.2$. The ROUGE-L score itself does not improve for $r {=} 0.2$, but note that this comes at the expense of more than doubling the output length. Also note that \textbf{L0-0.6} and \textbf{L0-0.8} only obtain relatively high scores because they produce long outputs that even exceed the length of the input. In fact, for $r = 0.6, 0.8$ no value of $\lambda_{len}$ produces outputs that are reasonably good ($> 10$ ROUGE-L) and short ($< 20$ BLEU) at the same time.

The above results confirm both our hypotheses: First (\hypothesisone), it is beneficial to use a BoV model over a single-vector model to reduce the compression issues induced by the fixed-size bottleneck. 
Secondly (\hypothesistwo), when using a BoV model, it is imperative to regularize the number of vectors in the bag as a way of smoothing the embedding space, making it easier to learn the mapping for unsupervised text generation tasks.
Moreover, if the number of vectors in the bag is large, our $\transformerplusplus$ architecture can substantially facilitate learning the mapping.

%% file: sections/qualitative-analysis.tex
\subsubsection{Qualitative Analysis}\label{app:yelp-sentences-qualitative}
We hypothesize that standard autoencoders suffer from poor performance with $\embtoemb$ if the text is too long to be encoded into a single vector (\hypothesisone). BoV-AEs were designed to alleviate this issue. Here, we conduct a qualitative analysis of 10 randomly selected model outputs on Yelp-Sentences. For comparability, we select models with similar levels of style transfer accuracy, namely the fixed size model with a performance of 59\% accuracy and 17 points self-BLEU to \textbf{L0-0.4} with a performance of 55\% accuracy and 38 points self-BLEU. We randomly sample 10 examples and show them in Table~\ref{tab:yelp-sentences-examples}. By design of the Yelp-Sentences dataset~\citep{shen2017style}, the inputs are sentences drawn from negative reviews, whose sentiment are supposed to be changed to positive. Note that due to how the dataset was constructed, some of the input sentences are already positive (\#7) or just neutral (\#2).

We observe several trends:
    \textbf{(1)} The fixed-sized model has a difficult time retaining the aspect discussed in the input sentence (\#10: staff instead of location, \#9: food instead of price), whereas the BoV-AE stays on topic. This is likely a consequence of the fixed-sized model's inability to encode the input well into a single vector, supporting \hypothesisone.
    \textbf{(2)} The outputs of the fixed-sized models are often completely unusable (\#1, \#2) or nonsensical (\#5, \#9, \#10), whereas the outputs of the BoV-AE are at least intelligible.
    \textbf{(3)} In absolute terms, the outputs of neither model are reliably grammatical or able to flip the sentiment. This is understandable since no large pretrained language model is used. 
    This would be needed to produce coherent outputs~\citep{brown2020language}, which then produces impressive outputs on style transfer~\citep{reif2021recipe}. As we argue in Section~\ref{sec:introduction}, our paper contributes to the foundation for large scale pretraining of autoencoder models to be used in $\embtoemb$.

\begin{table*}[]
    \centering
        \caption{10 randomly sampled examples from Yelp-Sentences and the outputs from each model.}
    \begin{tabularx}{\textwidth}{l|X||X|X}
        \# & \textbf{Input sentence} & \textbf{Output of fixed-size model} & \textbf{Output of L0-0.4} \\
        \hline
        \hline
        1 & generally speaking it was nothing worth coming back to . & but there here here and it will enjoy it . & generally remain it was it worth it and always happy ! \\
        \hline
        2&then why did n't they put some in ? & then she , you ta are the in the ? & then ' why n ' t they put some delicious !  \\
        \hline
        3&horrible experience ! & horrible ! & horrible experience ! \\
        \hline
        4&it was a shame because we were really looking forward to dining there . & it was a a fun , there and we have been to . & it really nice shame because we were really looking forward forward and fantastic ! \\
        \hline
        5&suffice to stay , this is not a great place to stay . & suffice to to not stay to this place is a stay . & suffice is not stay , this is a great place and always great !  \\
        \hline
        6&the chicken was weird . & the chicken was weird . & the chicken was weird . \\
        \hline
        7&my mom ordered the margarita panini which was pretty good . & my my margarita was ordered which was very good . & my mom ordered the margarita panini which was pretty good .  \\
        \hline
        8&i 'm not willing to take the chance . & i will definitely recommend your time or you . & i ' m not willing to take the great . \\
        \hline
        9&i would say for the price point that it was uninspired . & i had this place at the food , it 's super . & i would say for the price point that it was delicious . \\
        \hline
        10&the only pool complaint i have was from the last day of our stay . & the waitress was the the the the time here a last time  & the only pool complaint i have was from the day was wonderful ! 
    \end{tabularx}

    \label{tab:yelp-sentences-examples}
\end{table*}

%% file: sections/experiments_long.tex
\section{Experiments}

We test our model on unsupervised sentiment transfer: Given a negative review text, the goal is to rewrite it into a positive review while retaining as much of the original content as possible. However, there are no parallel input-output pairs given for training, we only have a corpus of positive and a corpus of negative reviews, respectively. Hence, this is an unsupervised task, a suitable test-bed for the Emb2Emb framework. Our evaluation consists of two parts. First, we evaluate on the sentence-wise Yelp dataset introduced by ~\cite{shen2017style} consisting of restaurant reviews. Examples in this dataset consist of single \emph{sentences} extracted from the reviews. Reconstruction of short texts like this can very well be learned by a fixed size vector model already. Therefore, we treat this dataset as a toy scenario in the sense that we use model sizes that are lower ($d = 32$) than today's standard. For computational reasons, we also use this setting to evaluate the different components of our model. In order to evaluate the benefits of our model in a realistic scenario, we construct a novel dataset whose examples consist of entire reviews with multiple sentences. These texts are considerably longer than those in the sentence-wise dataset, as Table~\ref{tab:dataset-stats} shows. 

\begin{table}[]
    \centering
    \begin{tabular}{c|c|c|c}
        Dataset & avg. \#words & \#pos & \#neg \\
        \hline
        Yelp-Sentences & $\approx 10$ & $\approx 230k$ & $\approx 230k$ \\
        Yelp-Reviews & $\approx 100$ & $500k$ & $500k$
    \end{tabular}
    \caption{Statistics for each dataset. \florian{Note: The numbers are rough estimates to prove the point, need to be updated with correct numbers.}}
    \label{tab:dataset-stats}
\end{table}

\paragraph{Evaluation metrics}
As noted above, in style transfer, two metrics are important: content retention and transfer ability. We measure the former with a separately trained classifier based on Distilbert, and content retention is measured in terms of self-BLEU. To obtain results at varying levels of transfer ability, we train the style transfer model with varying $\lambda_{sty}$.

Oftentimes, comparing models is easiest via a single value. Therefore, others have argued to aggregate content retention and transfer accuracy metrics~\cite{xu2018unpaired, krishna2020reformulatingunsupervised}. We follow the argumentation of \citet{krishna2020reformulatingunsupervised} that this aggregation should be per sentence:
\begin{equation}\label{eq:style-transfer-score}
    score = \frac{1}{M} \sum\limits_{i = 1}^{M} \operatorname{ACC}(s_y) \cdot \operatorname{BLEU}(s_y, s_x)
\end{equation}
where $s_x$ is the input sentence, $s_y$ is the predicted sentence, and $M$ is the number of data points. Note that BLEU here denotes sentencewise BLEU, not corpus-wide BLEU.

\subsection{Yelp-Sentences}\label{sec:exp-sentence-yelp}

\subsubsection{Experimental Setup}
In order to be able to show the usefulness of BoV-AEs on a dataset of short texts, we set the embedding size to $d = 32$.
We train all models using the Adam optimizer with an initial learning rate of $0.0001$. We train for for 1.5m steps and a batch size of 64 and save the model at the step with the lowest validation reconstruction loss. All models are trained with a denoising objective by dropping each word with a probability of 0.1.

\subsubsection{Reconstruction Ability}
The review texts in Yelp-Sentences are short, and if the latent vector has enough dimensions, can easily be reconstructed with a fixed-size bottleneck autoencoder. In that case, there is no benefit in using a BoV approach. Quite the contrary, it might rather be detrimental, because mapping between BoVs is harder than mapping a single vector.

\begin{figure}
    \centering
        \begin{minipage}{0.45\textwidth}
        \centering
        \includegraphics[width=\columnwidth]{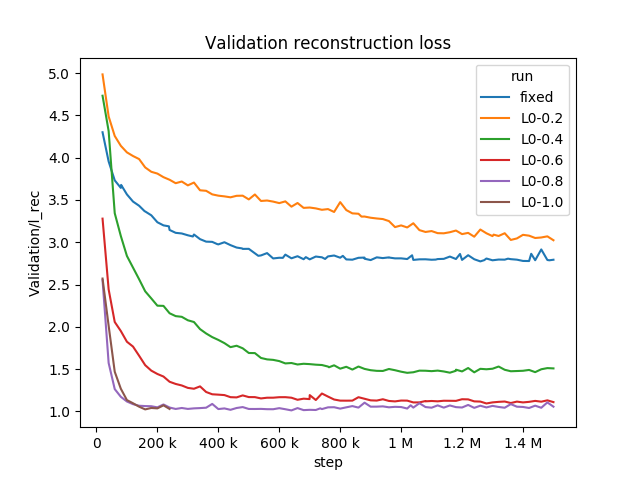}
        \caption{Reconstruction loss on the validation set for different autoencoders. \textbf{fixed}: The bag consists of a single vector obtained by averaging the embeddings at the last layer of the Transformer encoder. \textbf{L0-r}: BoV-AEs with L0Drop target ratio $r$.}
        
    \label{fig:val_rec_loss}
    \end{minipage}\hfill
    \begin{minipage}{0.45\textwidth}
        \centering
    \includegraphics[width=\columnwidth]{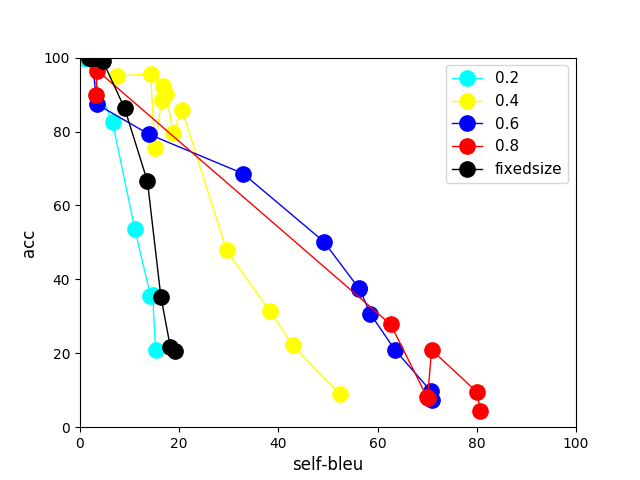}
        \caption{Style transfer performance of BoV models with different L0Drop target ratios compared to a fixed-size autoencoder. As both content retention and style transfer are important for style transfer, the further a graph is to the top right, the better the model.}
        \label{fig:sentence-yelp-l0drop}
    \end{minipage}
\end{figure}

However, when the latent vector is not large enough to well reconstruct the full sentence from a single vector, BoV provides a benefit. Figure~\ref{fig:val_rec_loss} demonstrates this, showing the reconstruction loss on the validation set for the fixed-size model compared to BoV models with different L0Drop target ratios $r$. 

The fixed-size autoencoder does not reach satisfactory reconstruction ability, converging at about 3. In contrast, BoV models are able to outperform the fixed-size model considerably, converging at a loss value of about 1 for $r = 0.8, 1.0$. As expected, higher target ratios lead to better reconstruction, because the model can use more vectors to store the information. In this setup, $r = 0.6$ seems to suffice, as the corresponding loss is very close to the optimal value. $r = 0.4$ yields to a higher loss, but is still drastically better than the fixed size model. $r = 0.2$ is not enough to outperform the fixed-size model.

\subsubsection{Style Transfer Ability}
To test the performance of BoV models on style transfer, we train a Transformer mapping ($\Phi = T$)\footnote{On the dataset with short texts, OffsetNet does not yield an advantage.} as described in Section~\ref{sec:emb2emb}. To obtain a results at varying transfer levels, we train multiple times with varying $\lambda_{sty}$. Further details can be found in the appendix~\ref{app:exp-details-sentence-yelp}.

Results are shown in Figure~\ref{fig:sentence-yelp-l0drop}. They correspond well to the reconstruction ability in that the BoV models with higher target ratios yield higher self-BLEU scores at comparable transfer abilities, topping off at $r = 0.6$. Again, the fixed-size model does considerably worse, only slightly beating the BoV model with $r = 0.2$.

\florian{From these experiments, it is not clear why we need L0Drop in the first place, and can't just use $r = 1.0$, i.e., not dropping any vectors, as the performance of $r = 1.0$ is similar in terms of automatic evaluation metrics. However, in preliminary experiments, I found that unregularized BoV-AEs ($r = 1.0$) produce nonsensical outputs that nonetheless fool the classifier. An example of this is shown in Table~\ref{tab:example}.}

\begin{table}[]
    \centering
    \begin{tabular}{c|c|c|c}
        Input & $r = 0.4$ & $r = 0.8$ & $r = 1.0$ \\
        \hline
        \makecell{the food \\ is terrible} & \makecell{the food \\ is great} & \makecell{great food \\ is terrible} & \makecell{great food \\ is terrible}
    \end{tabular}
    \caption{Example output of BoV-AE with various target ratios $r$. BoV-AEs with no or mild regularization produce non-sensical outputs (which nonetheless fool the classifier), whereas stronger regularization produces meaningful outputs.\florian{Warning: This is a cherry-picked example that I recalled from memory; in the final paper, it should probably be replaced with a randomly sampled list of examples.}}
    \label{tab:example}
\end{table}

\subsubsection{Ablations}

\paragraph{Window size}

The window size $k$ determines which bag sizes around the input bag size we backpropagate from. Here, we investigates its influence on the model's performance. Since the $\lambda_{sty}$ hyperparameter is very sensitive to other model hyperparameters, we train with varying $\lambda_{sty}$ for each fixed window size and report the best style transfer score (Eq. \ref{eq:style-transfer-score} for each window size. Further experimental details can be found in appendix~\ref{app:exp-details-sentence-yelp}.

Figure~\ref{fig:exp-window-size} shows that increasing the window size is beneficial up to some point ($k = 5$), whereas increasing by too much ($k = 20$) is detrimental to model performance even compared to a window size of zero. We hypothesize that backpropagating bags that are either very small or very large is detrimental because it forces the model to adjust its parameters to optimize unrealistic bags, taking away capacity for fitting realistic bags.
\begin{figure}
    \centering
        \begin{minipage}{0.45\textwidth}
        \centering
        \includegraphics[width=\columnwidth]{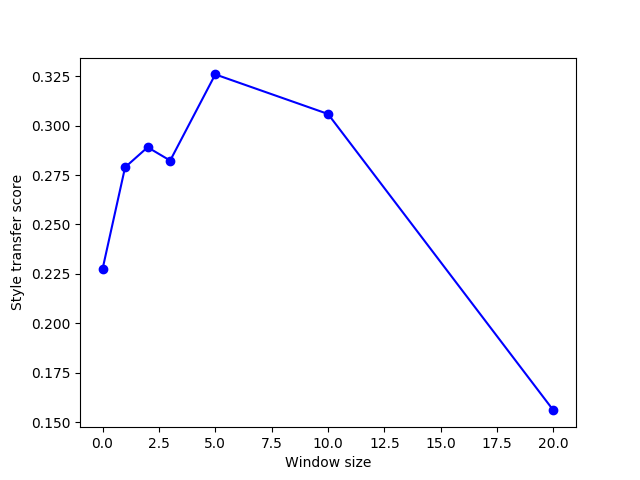}
        \caption{Style transfer score depending on the window size.}
    \label{fig:exp-window-size}
    \end{minipage}\hfill
    \begin{minipage}{0.45\textwidth}
        \centering
        \includegraphics[width=\columnwidth]{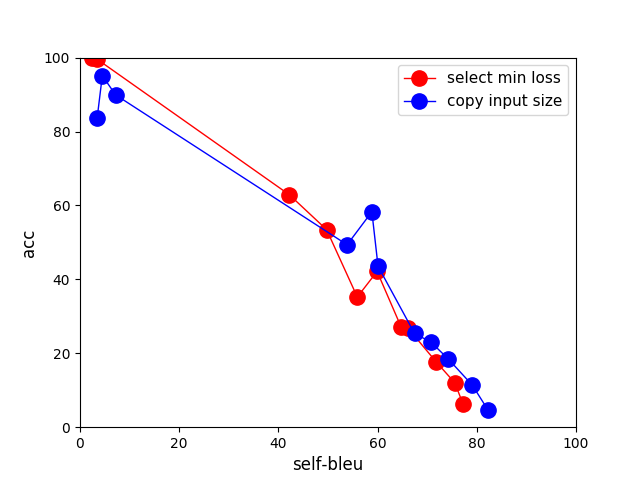}
        \caption{Comparison of the two output bag size selection methods.}
   \label{fig:input-size-selection}
    \end{minipage}

\end{figure}

\paragraph{Differentiable Hausdorff}
In Section~\ref{sec:hausdorff-sim}, we argue that the min operation should be replaced by softmin in order to facilitate backpropagation. Here, we test if the differentiable version is really necessary, that is, we compare Eq.~\ref{eq:avg-hausdorff} to Eq.~\ref{eq:diff-hausdorff}. Like before, we train the two variants with different $\lambda_{sty}$ and select the best style transfer score on the validation set. As always, further experimental details can be found in appendix~\ref{app:exp-details-sentence-yelp}.

The difference is substantial: Average Hausdorff reaches 0.146, whereas differentiable Hausdorff reaches 0.242. We hypothesize that this discrepancy is due to the difficult nature of the style transfer problem, which requires carefully balancing the two objectives content retention (via Hausdorff) and style transfer (via the classifier). This is easier when the objective functions are smooth, which is the key advantage of differentiable Hausdorff.

\paragraph{Output bag selection}
In sentiment transfer, we know that the output bag should be approximately as large as the input bag. However, as we argued in Section~\ref{sec:method-variable-size-bag}, this is not the case in all tasks. For these cases, we proposed to choose the bag size that achieves the minimal loss. Here, we investigate how well this method works by comparing it to a model that always chooses the input bag size.

In Figure~\ref{fig:input-size-selection} we observe that selecting the bag size based on minimum loss is slightly worse than just setting the output size equal to the input size, but gets reasonably close. This leads us to believe that our model will also work on tasks for which no good prior on the output size exists.

\subsection{Yelp-Reviews}

\subsubsection{Dataset}
In order to demonstrate the usefulness of our model on long texts, we turn to the original Yelp dataset\footnote{The dataset was obtained from \url{https://www.yelp.com/dataset} in May 2021.}. Our goal is to obtain texts long enough such they cannot be reconstructed by a reasonably sized autoencoder with a single-vector bottleneck. We find that to be the case when limiting ourselves to reviews of maximum 100 words\footnote{We apply this limit due to the computational complexity of Transformers on long texts.}. Otherwise, we stick with similar filtering criteria as \citet{shen2017style}: We only consider restaurant businesses. We consider reviews with 1 or 2 stars as negative, and reviews with 5 stars as positive. We don't consider reviews with 3 or 4 stars to avoid including neutral reviews. We subsample 400,000 positive and negative reviews for training, respectively, and use 50,000 for validation and test set each.

\subsubsection{Experimental Setup}
We train all models with vectors of size $d = 512$, which is a reasonable size comparable to recent models~\cite{shen2019latent}. The autoencoder's encoder and decoder have 3 layers each and are trained with Adam (lr=0.0001) for 500k steps. We save the model at the step with the lowest validation reconstruction loss. All models are trained with denoising by dropping each word with a probability of 0.1.

\subsubsection{Reconstruction Ability}
In Section~\ref{sec:exp-sentence-yelp}, we presented evidence that single-vector autoencoders cannot learn to reliably reconstruct short texts if the vector size is not large enough. In Figure~\ref{fig:review-yelp-l0}, we investigate the same question for longer texts (cmp. Table~\ref{tab:dataset-stats}).

Again, despite a large dimensionality ($d = 512$), the single-vector model achieves substantially lower reconstruction ability than BoV-AE. With respect to the target sparsity rate, we find that $r = 0.4$ is enough to reach dramatically better results than the fixed-size model, whereas $r = 0.2$ does not suffice.

\begin{figure}
    \centering
    \includegraphics[width=\columnwidth]{images/reconstruction_reviewyelp.png}
    \caption{Reconstruction loss on the validation set for different autoencoders. \textbf{fixed}: The bag consists of a single vector obtained by averaging the embeddings at the last layer of the Transformer encoder. \textbf{L0-r}: BoV-AE with L0Drop target ratio $r$.}
    \label{fig:review-yelp-l0}
\end{figure}

\subsubsection{Style Transfer Ability}
\florian{We need convincing results here, which we don't have yet. Results of intermediate experiments are shown in Table~\ref{tab:yelp-full-prelim-results}.
}

\begin{itemize}
    \item If trained from scratch, both the standard Transformer mapping and the OffsetNet adaptation achieve results with self-BLEU scores essentially being zero.
    \item If we pretrain the OffsetNet adaptation on a denoising objective, it attains around 2 self-BLEU points at a reasonable level of sentiment transfer ability, but it is still worse than the single-vector model.
\end{itemize}

\begin{table}[]
    \centering
    \begin{tabular}{c|c|c}
         Model & self-BLEU & Acc. \\
         \hline
        fixed & $\approx 2$ & $\approx 90$ \\
        \hline
        L0-0.6 + $\Phi_{T}$ &$\approx 0$ & $\approx 90$ \\
        \hline
        L0-0.6 + $\Phi_{Offset}$ &$\approx 0$ & $\approx 90$ \\
        \hline
        \makecell{L0-0.6 + Denoising \\ Pretraining + $\Phi_{Offset}$} &$\approx 2$ & $\approx 60$
    \end{tabular}
    \caption{Preliminary experimental results on Yelp-Reviews. $\Phi_{Offset}$: The mapping is a standard transformer decoder. $\Phi_{Offset}$: The mapping is the copy-based OffsetNet adaptation. None of the models work well on the dataset with full reviews.}
    \label{tab:yelp-full-prelim-results}
\end{table}

%% file: sections/related.tex
\section{Related Work}\label{sec:problem}

\ifthenelse{\boolean{short}}
{}
{
\paragraph{Hausdorff Distance}
The Hausdorff distance is often used as a performance evaluation metric, e.g., in medical image segmentation~\cite{taha2015metrics, aydin2020usage} or multi-objective optimization problems~\cite{6151115, marti2015averaged}. Besides that, it is also a relatively common approach in the vision domain as a means of comparing images~\cite{huttenlocher1993comparing}, e.g., for face recognition~\cite{takacs1998comparing, lin2003spatially}, and word image recognition~\cite{lu2001approach}.
More recently, variants of the Hausdorff metric have also been used as loss functions to train neural networks: For point set prediction in 3D object reconstruction~\cite{fan2017point}, for locating objects~\cite{ribera2019locating}, and for vehicle reidentification~\cite{zhao2021phdlearninglearning}.

In NLP, Hausdorff-like metrics have rarely been used. \citet{chen2019learning} uses the metric to measure distances of embeddings from different languages in multilingual neural models. \citet{nutanong2016scalable} view authorship attribution as a set similarity problem, and use some variants of the Hausdorff distance for measuring set similarity. Recently, \citet{kuo2020compositional} presented a model for grounded language learning in which the Hausdorff distance of two attention maps is maximized. To the best of our knowledge, our paper is the first to present a novel, fully differentiable variant of the Hausdorff distance as a loss for language learning.
}

\ifthenelse{\boolean{short}}{\textbf{Manipulations in latent space:}}{\paragraph{Manipulations in Latent Space}}
Besides $\embtoemb$, latent space manipulations for textual style transfer are performed either via gradient descent~\citep{wang2019controllable, liu2019revision} or by adding constant style vectors to the input~\citep{shen2019latent, montero2021sentence}.
In computer vision, discovering latent space manipulations for image style transfer has recently become a topic of increased interest, in both supervised~\citep{jahanian2019steerability, zhuang2021enjoy} and unsupervised ways~\citep{harkonen2020ganspace, voynov2020unsupervised}.
While these vision methods are similar to $\embtoemb$ conceptually, they differ from our work in important ways.
First, they focus on the latent space of GANs~\citep{goodfellow2014generative}, which work well for image generation but are known to struggle with text~\citep{caccia2018language}.
Secondly, images typically have a fixed size, and consequently their latent representations consist of single vectors.
Our work focuses on data of variable size, which may have important insights for modalities other than text, e.g. videos and speech.

\ifthenelse{\boolean{short}}{\textbf{Unsupervised conditional text generation:}}{\paragraph{Unsupervised Conditional Text Generation}}
\ifthenelse{\boolean{short}}
{
Modern unsupervised conditional text generation approaches are based on either \textbf{(a)} language models (LMs) or \textbf{(b)} autoencoders (AEs). \textbf{(a)} One type of LM approach explicitly conditions on attributes during pretraining~\citep{keskar2019ctrl}, which puts restrictions on the data that can be used for training. Another type adapts pretrained LMs for conditional text generation by learning modifications in the embedding space~\citep{dathathri2019plug}.
These approaches work well because LMs are pretrained with very large amounts of data and compute power, which results in exceptional generative ability~\citep{radford2019language, brown2020language} that even enables impressive zero-shot style transfer results~\citep{reif2021recipe}.
However, in contrast to AEs, LMs are not designed to have a latent space that facilitates learning in it. 
We therefore argue that AE approaches could perform even better than LMs if they were given equal resources. This motivates our research.
\textbf{(b)} A very common approach to AE-based unsupervised conditional text generation is to learn a shared latent space for input and output corpora that is agnostic to the attribute of interest (e.g., sentiment transfer~\citep{shen2017style}, style transfer~\citep{subramanian2018multiple}, summarization~\citep{liu2019summae}, machine translation~\citep{artetxe2017unsupervised}). However, in these approaches, the decoder is explicitly conditioned on the desired attribute that must be available for all data points, complicating pretraining on unlabeled data. 
To overcome this, \citet{mai2020plug} recently proposed $\embtoemb$, which disentangles AE pretraining from learning to change the attributes via a simple mapping. Our paper makes an important contribution by improving the expressivity of $\embtoemb$ through variable-size representations.
}
{The most common approach to text style transfer is to learn a disentangled shared latent space that is agnostic to the style of the input. Style transfer is then achieved by training the decoder conditioned on the desired style attribute,  ~\cite{hu2017toward, shen2017style, fu2018style, zhao2018adversarially, subramanian2018multiple, li2018delete, logeswaran2018content, yang2018unsupervised, li2019domain}, which hinders their employment in a plug and play fashion.
Most methods either rely on adversarial objectives~\cite{shen2017style, hu2017toward, fu2018style}, retrieval~\cite{li2018delete}, or backtranslation~\cite{subramanian2018multiple, logeswaran2018content} to make the latent codes independent of the style attribute.
Notable exceptions are Transformer-based~\cite{dai2019style, sudhakar-etal-2019-transforming}, use reinforcement learning for backtranslating through the discrete space~\cite{liu2019transformer}, build pseudo-parallel corpora~\cite{kruengkrai2019learning, jin2019imat}, or modify the latent-variable at inference time by following the gradient of a style classifier~\cite{wang2019controllable,liu2019revision}.
Similar to our motivation, %
 \cite{li2019domain} aim at improving in-domain performance by incorporating out-of-domain data into training. However, because their model again conditions on the target data, they have to train the AE jointly with the target corpus, defeating the purpose of large-scale pretraining.

In contrast to previous methods, Emb2Emb can be combined with any pretrained AE even if it was not trained with target attributes in mind.
It is therefore very close in spirit to plug and play language models %
by \cite{dathathri2019plug} who showed how to use pretrained language models for controlled generation without any attribute conditioning (hence, the name). It is also similar to pretrain-and-plugin variational AEs~\cite{duan2019pre}, who learn small adapters with few parameters for a pretrained VAE to generate latent codes that decode into text with a specific attribute. 
However, these models cannot be conditioned on input text, and are thus not applicable to style transfer. It is therefore close in spirit to plug and play language models by \cite{dathathri2019plug} and pretrain-and-plugin variational AEs by \cite{duan2019pre}, but conditions on an input text \emph{and} a target attribute, rather than \emph{only} the target attribute.

}
\iffalse
\florian{TODO: Style transfer on vision}

\paragraph{Textual Autoencoders}
 \florian{TODO: blend with variable size autoencoders}
\ifthenelse{\boolean{short}}
{}
{
Autoencoders are a very active field of research, %
leading to constant progress through denoising~\cite{vincent2010stacked}, variational~\cite{kingma2013autoencoding, Higgins2017betaVAELB, dai2019diagnosing}, adversarial~\cite{makhzani2015adversarial, zhao2018adversarially}, and, more recently, regularized~\cite{Ghosh2020From} autoencoders, to name a few.
Ever since \cite{bowman2015generating} adopted variational autoencoders for sentences by employing a recurrent sequence-to-sequence model, %
 improving both the architecture~\cite{severyn2017hybrid, prato2019towards, liu2019transformer, gagnon2019salsa} and the training objective~\cite{zhao2018adversarially, shen2019latent} have received considerable attention. The goal is typically to improve both the reconstruction and generation performance~\cite{cifka2018eval}.

Our framework is completely agnostic to the type of autoencoder that is used, as long as it is trained to reconstruct the input. Hence, our framework directly benefits from any kind of modelling advancement in autoencoder research.

 }
 \fi

%% file: sections/conclusion_short.tex
\section{Conclusion}
Our paper addresses a fundamental research question: How do we learn text representations in such a way that conditional text generation can be learned in the latent space (e.g. $\embtoemb$)? We propose Bag-of-Vectors Autoencoders to overcome the fundamental bottleneck of single-vector autoencoders:
Controlled experiments revealed that, thanks to our technical contributions, BoV-AEs perform substantially better at learning in their embedding space when the text is too long to be encoded into a single vector.
This lays the foundation for learning conditional long-text generation models in a framework such as $\embtoemb$ in an unsupervised manner.
\ifthenelse{\boolean{short}}{}{A shortcoming of the model is that they are more sensitive with respect to hyperparameters when trained on the downstream task, destabilizing training. However, while it can be difficult to train models to their optimum (as is typical in deep learning), BoV-AEs are still better than the fixed-sized baseline.}

\short{}{
The state-of-the-art in practically all language-related tasks relies heavily on large-scale pretraining, which requires large amounts of resources. Our study is fundamental in nature; we systematically demonstrate the benefit of $\embtoemb$ with variable-size representations rather than fixed-sized representations via controlled experiments.
Nevertheless, our model is in principle fit for the future. A unique advantage of the $\embtoemb$ framework is its compatibility with pretrained autoencoders. \citet{mai2020plug} showed that the $\embtoemb$ framework benefits immensely from unlabeled data. 
Moreover, in Appendix~\ref{app:yelp-sentences-pretraining}, we discuss promising results of an initial study that makes the pretrained autoencoder BART~\citep{lewis2019bart} compatible with $\embtoemb$ by further finetuning it with L0Drop regularization.
This indicates that, given enough compute and data for large-scale pretraining from scratch, Bag-of-Vectors Autoencoders could have the potential to become a \emph{Foundation Model}~\citep{bommasani2021opportunities} like BERT, BART, and GPT-3.
Our study paves the way for the application of BoV-AEs for unsupervised tasks by demonstrating how to learn in their latent space.
}

\section*{Acknowledgements}
Florian Mai was supported by
the Swiss National Science Foundation under grant
number 200021\_178862.

%% file: sections/ethics.tex
\paragraph{Applications}
The focus of our study is not any particular application, but concerns fundamental questions in unsupervised conditional text generation in general.
Unsupervised applications are useful in scenarios where few annotations exists, which is particularly common in understudied low-resource languages (e.g. unsupervised neural machine translation~\citep{kuwanto2021low}).
Of course, oftentimes unsupervised solutions perform worse than supervised ones, requiring extra care during deployment to avoid harm from potential mistakes.

Despite the fundamental nature of our study, we test our model on two concrete problems, \textbf{a)} text style transfer and \textbf{b)} sentence summarization.
\textbf{a)} Style transfer has applications that are beneficial to society, such as expressing "complicated" text in simpler terms (\emph{text simplification}) or avoiding potentially offensive language (\emph{detoxification}), both of which are particularly beneficial for traditionally underprivileged groups such as non-native English speakers.
However, the same technology can also be used maliciously by simply inverting the style transfer direction.
In this paper, we decided to study sentiment transfer of restaurant reviews as a style transfer task. 
The reasons are primarily practical; deriving both from the Yelp dataset, we can study the effectiveness of our model on two datasets (sentences and full reviews) that are very similar in content but considerably different in length.
On one hand, this allows us to demonstrate the effectiveness of our model in a realistic, but computationally demanding setting. On the other hand, we can perform ablations in a less expensive setting.
Apart from serving as a test bed for scientific research, sentiment transfer itself has no obvious real-world application. With enough imagination one can construe a scenario where a bad actor hacks into the database of a review platform like Yelp to e.g. manipulate the content of existing reviews.
However, we rate this as highly unrealistic due to high opportunity cost, as it is much easier to generate fake reviews with large language models rather than hack into a system and alter existing reviews.

\textbf{b)} Summarization systems can be very valuable for society by enabling people to process information faster. But this depends on the system's output to be mostly factual, which neural summarization systems struggle with~\citep{maynez2020faithfulness}.
Unfaithful outputs may convey misinformation, which can potentially harm users.

\paragraph{Deployment}
While we argue above that sentiment transfer has no useful real-world application, the model can still be deployed for demonstration purposes, or be trained and deployed for other tasks, e.g., sentence simplification. However, we urge not to deploy the models developed in this paper directly without adaptation for several reasons. i) The absolute performance is suboptimal (e.g., no large-scale pretraining) and hence makes many mistakes that a real-world application should avoid to prevent harm. ii) The model can occasionally produce toxic output. Of course, the extent to which this happens strongly depends on the training data. E.g., Yelp restaurant reviews can sometimes contain vulgar language. Any real-world application should hence consider pre- and post-filtering methods. iii) The model might be biased towards certain populations, the extent of which is not the subject of this study. For example, the sentiment transfer models would likely work better for fast food restaurants than restaurants of African cuisine, because the former is more common in the mostly US-centric data that the model is trained on. A real-world application needs to consider the requirements of the target audience.

Similarly, we argue that the sentence summarization model studied in this paper needs further improvements before deployment, some of which we mentioned in the main paper. Large-scale pretraining could also help to mitigate hallucinated facts~\citep{maynez2020faithfulness}.
\paragraph{Dataset}
The Yelp-Reviews dataset is a direct derivative of the Yelp Open Dataset\footnote{\url{www.yelp.com/dataset}}. Their license agreement states that any derivative remains the property of Yelp, hence we can not directly release the dataset.
However, academic researchers can easily obtain their own license for non-commercial use and recreate the dataset used in this study via the script we provide in the supplementary material.
No further data collection was conducted.

We explicitly try to avoid the inclusion of sensitive data (e.g., the name of a Yelp reviewer) for training and evaluation by only using the review text and no attached meta-data.

%% file: sections/limitations.tex
\section*{Limitations}
Our study is fundamental in nature; we systematically demonstrate the benefit of $\embtoemb$ with variable-size representations rather than fixed-sized representations via controlled experiments. We do not aim to maximize the performance on any specific task. This implicates some limitations.

\paragraph{Applications}
We discourage application engineers to apply our model without modification in production for text style transfer or unsupervised summarization.

First, the state-of-the-art in practically all language-related tasks relies heavily on large-scale pretraining, which requires large amounts of resources. For example, the state-of-the-art in text style transfer by \citet{reif2021recipe} is built upon a language model with 137B parameters~\citep{thoppilan2022lamda}. Due to this foundation, the model is able to generalize to arbitrary text style transfer tasks in a zero-shot manner, generating far better outputs than our models. The best unsupervised text summarization models also require large language models~\citep{brown2020language}.
Second, we abstain from task-specific tweaks to our model such as backtranslation for style transfer~\citep{lample2018unsupervised}.

However, we view both these factors as orthogonal to our contribution. Our model is in principle compatible with large-scale pretraining. In fact, a unique advantage of the $\embtoemb$ framework is its compatibility with pretrained autoencoders. \citet{mai2020plug} showed that the $\embtoemb$ framework, a state-of-the-art model for text style transfer before pretrained models became ubiquitous, benefits immensely from unlabeled data.
Moreover, in Appendix~\ref{app:yelp-sentences-pretraining}, we discuss promising results of an initial study that makes the pretrained autoencoder BART~\citep{lewis2019bart} compatible with $\embtoemb$ by further finetuning it with L0Drop regularization.
The resulting model produces more fluent and grammatical outputs than the model trained from scratch.
This indicates that, given enough compute and data for large-scale pretraining from scratch, Bag-of-Vectors Autoencoders could have the potential to become a \emph{Foundation Model}~\citep{bommasani2021opportunities} like BERT, BART, and GPT-3.
Our study paves the way for the application of BoV-AEs for unsupervised tasks by demonstrating how to learn in their latent space.

\paragraph{Hyperparameter sensitivity}
BoV-AEs are more sensitive with respect to certain hyperparameters than their fixed-sized counterparts. We noticed this in two places. First, when pretraining on unlabeled data, BoVAEs required a more finegrained learning rate than fixed-sized AEs.
This is also notable whne comparing their learning curves: The curves in Figure~\ref{fig:val_rec_loss} are smoother than in Figure~\ref{fig:yelp-reviews-valrec}.
Secondly, the tradeoff between content retention and transfer ability is not as easily controllable through the $\lambda_{sty}$  hyperparameter as in the fixed-sized model.
For instance, in Figure~\ref{fig:exp-yelp-reviews-downstream}, the Pareto front of the fixed-sized model is considerably smoother.
However, while it can be difficult to train models to their optimum (as is typical in deep learning), BoV-AEs can still drastically outperform the fixed-sized baseline.
Nonetheless, for practical purposes it will be important to discover more robust hyperparameterization similar to Equation~\ref{eq:l0drop}.

\paragraph{Computation time}
BoV-AEs are more sophisticated than standard fixed-size AEs, and this also comes with higher computational cost.
We analyze this in depth in Appendix~\ref{app:yelp-sentences-time}. In summary, especially the mapping is considerably more costly, as it depends on the input length. However, this cost is mitigated through L0Drop's sparsification, and for very long texts, fixed-size AEs are no viable option.
Nonetheless, investigating the suitability of efficient Transformer alternatives for our framework will be an important future research avenue.

%% file: sections/reproducibility.tex
We took several precautions to ensure that our work is reproducible.

\paragraph{Datasets}
Our study is based on two existing datasets, Gigaword sentence summarization, and Yelp-Sentences style transfer. For these two datasets, we provide scripts that preprocess them as in our study. 
For Yelp-Reviews dataset, we provide a detailed description in appendix~\ref{sec:app-yelp-reviews-dataset}. Moreover, we provide a script that allows to construct the dataset as a derivative from Yelp data. 
In order to get access to Yelp data, practitioners have to obtain a license from Yelp that is free of charge. The data may only be used for non-commercial or academic purposes, but this suffices to reproduce our study.
The Gigaword corpus is commonly used, and can be downloaded from the Linguistic Dataset Consortium at \url{https://catalog.ldc.upenn.edu/LDC2012T21}.
For downloading, a membership is mandatory, or otherwise fees apply. However, this commonplace in NLP research institutes.

\paragraph{Code}
We provide code to reproduce all our experiment in the supplementary materials.

\paragraph{Experiments}
We provide details on each experiment's setup in the appendix.
However, it's impractical to report all details that may impact the outcome.
Therefore, for each experiment we additionally provide a csv file in the supplementary material. The file contains information on all training parameters, model hyperparameters and results. 
In combination with the code, this allows to reconstruct almost the exact experimental setup used in our study apart from parameters that are beyond our control, such as the computation environment.

%% file: sections/appendix.tex
\section{Sentence Summarization}\label{app:sentence-summarization}
We perform experiments on unsupervised sentence summarization~\citep{rush2015neural} for two main reasons.
First, we would like to understand whether our conclusions hold for more tasks than just text style transfer.
Second, the sentence summarization dataset consists of texts of medium length, between the length of Yelp-Review and Yelp-Sentences.
This length is long enough to showcase the benefit of $\transformerplusplus$, yet still computationally cheap enough to conduct this expensive ablation study.

\input{sections/sentence-summarization-eval}
\input{sections/sentence-summarization}
\section{Experimental Details}\label{app:exp-details}
Here, we describe the experimental setup used in our experiments. We try to be exhaustive, but the exact training configurations and code will also be given as downloadable source code for reference.

\begin{table*}
    \centering
        \caption{Basic statistics for each dataset used in this study. Average number of words refers to input texts and output texts, respectively.}
    \begin{tabular}{c|c|c|c}
        Dataset & avg. \#words & \#inputs & \#outputs \\
        \hline
        Yelp-Sentences & 9.7 / 8.5 & 177k & 267k \\
        Gigaword & 27.2 / 8.2 & 500k & 500k \\
        Yelp-Reviews & 56.1 / 48.7 & $500k$ & $500k$ 
    \end{tabular}
    \label{tab:dataset-stats}
\end{table*}

\subsection{Implementation} We implemented BoV-AEs and fixed-sized AEs within the codebase. Neural networks are implemented via PyTorch~\citep{paszke2019pytorch}. The code is provided with the supplementary material, and will be makde available publicly under the MIT license when the paper is published. 
For each dataset, we train a new BPE tokenizer~\citep{sennrich2015neural} via Huggingface tokenizer library~\citep{wolf2019huggingface}. We limit the vocabulary to the 30k most frequent tokens.
We use NLTK~\citep{bird2009natural} for computing sentence-wise BLEU scores and a Python-based reimplementation of ROUGE-1.5.5. for all  ROUGE scores\footnote{\url{https://pypi.org/project/rouge-score/}}.
We run our experiments on single GPUs, which are available to us as part of a computation grid.
Specific GPU assignment is outside of our control, and the specific GPUs vary between GeForce GTX Titan X and RTX 3090 in power.

We estimate the total compuational cost of the experiments reported in this paper to be 7530 GPU hours.
The majority of this cost is on autoencoder pretraining, which accounts for 6640h (cmp. 890h for downstream training). Due to the long inputs and relatively large models, pretraining on Yelp-Reviews is by far the most costly (5760h).

Note that a sufficiently large and generic model has to be pretrained only once and could be applied to a wide range of downstream tasks, as is the case for e.g. BERT. In our experiments, we had to pretrain on each corpus separately. 

We estimate the computational budget over the whole development stage of this study to be around 25,000 GPU hours.

\subsection{Autoencoder Pretraining}\label{app:experimental-details-ae}
All autoencoders consist of standard Transformer encoders and decoders~\citep{vaswani2017attention}, with 3 encoder and decoder layers, respectively. The Transformers have 2 heads and the dimensionality is set to the same as the latent vectors (Yelp-Reviews: 512, Yelp-Sentences: 32, Gigaword: 128). The total number of parameters of each model is shown in Table~\ref{tab:num-parameters}. BoV-AEs are marginally larger due to the L0Drop layers. In case of the fixed sized model, the representations at the last layer are averaged. Otherwise we perform L0Drop as described in Section~\ref{sec:bovae}. We set $\lambda_{L_0} = 10$ for all BoV models and only vary the target ratio. All models are trained with a dropout~\citep{DBLP:journals/jmlr/SrivastavaHKSS14} probability of $0.1$ and a denoising objective, i.e, tokens have a chance of 10\% to be dropped from the sentence. We train the model with the Adam optimizer~\citep{kingma2014adam} with an initial learning rate of $lr = 0.00005$ (Yelp-Reviews and Gigaword) or $lr = 0.0001$ (Yelp-Sentences) and a batch size of 64. We experimented with other learning rates ($0.00005, 0.0005$) for the fixed-size model on Yelp-Reviews, but the results did not improve. Models are trained for 2 million steps on Gigaword and Yelp-Reviews and for 1.5 million steps on Yelp-Sentences. We check the validation set performance every 20,000 steps and select the best model according to validation reconstruction performance.

All the above hyperparameters were set once and not changed during the development of BoV-AEs, except for the learning rate of Adam. BoV-AE in particular is sensitive to this hyperparameter on the Yelp-Review dataset. We hence conducted a small grid search on $lr \in \{0.0005, 0.0002, 0.0001, 0.00005\}$ for \textbf{L0-0.2} to determine the best value reported above. We then used that same learning rate to all other configurations on Yelp-Reviews.
\begin{table*}[]
    \centering
    \begin{tabular}{c|c|c|c}
    & Yelp-Reviews & Yelp-Sentences & Gigaword \\
    \hline
        Fixed-size AE & 14.578m & 0.958m & 2.758m \\
        BoV-AE & 15.1m & 0.960m & 2.725m
    \end{tabular}
    \caption{Number of parameters of pretrained autoencoders.}
    \label{tab:num-parameters}
\end{table*}

\subsection{Downstream Task Training}
After the autoencoder pretraining, we train downstream by freezing the parameters of the encoder and decoder. The dimensionality of the one-layer mapping $\Phi$ (a Transformer decoder with 4 heads) is set to the same as the latent representation (Yelp-Reviews: 512, Yelp-Sentences: 32, Gigaword: 128). We set the maximum number of output vectors to $N = 250$ on Yelp-Reviews and Gigaword, and $N = 30$ on Yelp-Sentences. The batch size is 64 for Yelp-Sentences and Gigaword and 16 on Yelp-Reviews. We train for 10 epochs on Yelp-Sentences and Gigaword, and for 3 epochs on Yelp-Reviews. The validation performance is evaluated after each epoch.

\textbf{Losses:} In all tasks we have two loss components. For $\loss_{sim}$, we use differentiable Hausdorff unless specified otherwise (in the ablation).
$\loss_{sty}$ and $\loss_{len}$ depend on classifiers / regressors, which we train separately after the autoencoder pretraining as a one-layer Transformer encoder. The embeddings are then averaged and plugged into a one-layer MLP whose hidden size is half of the input size and uses the $\tanh$ activation function. These classifiers are trained via Adam ($lr = 0.0001$) for 10 epochs and we evaluate the validation set performance after each. The total loss depends on a window size as described in Equation~\ref{eq:windowsize}. For performance reasons (multiple computations of the loss), we set $k = 0$ unless specified differently.

\subsection{Yelp-Reviews}

\subsubsection{Dataset}\label{sec:app-yelp-reviews-dataset}
The dataset was obtained from \url{https://www.yelp.com/dataset} in May 2021. Our goal is to obtain texts long enough such they cannot be reconstructed by a reasonably sized autoencoder with a single-vector bottleneck. We find that to be the case when limiting ourselves to reviews of maximum 100 words. We apply this limit due to the computational complexity of Transformers on long texts. Otherwise, we stick with similar filtering criteria as \citet{shen2017style}: We only consider restaurant businesses. We consider reviews with 1 or 2 stars as negative, and reviews with 5 stars as positive. We don't consider reviews with 3 or 4 stars to avoid including neutral reviews. We subsample 400,000 positive and negative reviews for training, respectively, and use 50,000 for validation and test set each.

In order to demonstrate the usefulness of our model on long texts, we turn to the original Yelp dataset\footnote{The dataset was obtained from \url{https://www.yelp.com/dataset} in May 2021.}. Our goal is to obtain texts long enough such they cannot be reconstructed by a reasonably sized autoencoder with a single-vector bottleneck. We find that to be the case when limiting ourselves to reviews of maximum 100 words\footnote{We apply this limit due to the computational complexity of Transformers on long texts.}. Otherwise, we stick with similar filtering criteria as \citet{shen2017style}: We only consider restaurant businesses. We consider reviews with 1 or 2 stars as negative, and reviews with 5 stars as positive. We don't consider reviews with 3 or 4 stars to avoid including neutral reviews. We subsample 400,000 positive and negative reviews for training, respectively, and use 50,000 for validation and test set each.

\subsubsection{Downstream Training}\label{app:exp-details-review-yelp}
For both the fixed-size model and the BoV model (\textbf{L0-0.1}), we choose the best learning rate among $lr = 0.0001$ and $lr = 0.0005$ on the validation set and report test set results. We train with $\loss_{sty} \in \{0.1, 0.2, 0.5, 1, 2, 5, 10\}$, resulting in the scatter plot in Figure~\ref{fig:exp-yelp-reviews-downstream}.

\subsection{Yelp-Sentences}

\subsubsection{Dataset}
Yelp-Sentences consists of the sentiment transfer dataset created by \citet{shen2017style}, who made their data available at \url{https://github.com/shentianxiao/language-style-transfer/tree/master/data/yelp}. We use their data as is without further preprocessing. Table~\ref{tab:dataset-stats} presents some basic statistics about this dataset.

\subsubsection{Downstream Training}
\label{app:exp-details-sentence-yelp}

We train BoV models with $\lambda_{sty} \in \{1, 2, 5, 10, 20, 50, 100\}$. To make sure that our results are not due to insufficient tuning, for the fixed-sized model, we use the following larger range: $\{0.01, 0.02, 0.05, 0.1, 0.2, 0.5, 1, 2, 5, 10, 20, 50, 100 \}$. All configurations are trained with $lr = 0.0005$. These results produce the scatter plot in Figure~\ref{fig:sentence-yelp-l0drop}.

\subsubsection{Ablations}\label{app:exp-details-sentence-yelp-ablations}
For the ablations on differentiable Hausdorff distance and the window size, we use the \textbf{L0-0.6} model. For each option, we train with  $\loss_{sty} \in \{0.1, 0.2, 0.5, 1, 2, 5, 10, 20, 40, 60, 80, 100 \}$ and report the best value in terms of style transfer score on the validation set.

\subsection{Sentence Summarization}

\subsubsection{Dataset}
The dataset is based on the Gigaword corpus~\citep{graff2003english}. We largely follow the preprocessing in ~\cite{rush2015neural}, which we obtained from the paper's GitHub repository at \url{https://github.com/facebookarchive/NAMAS}. Different from them, we convert all inputs and outputs to lower case and use a smaller split (1 million examples).
We provide the scripts for constructing the dataset from a copy of the Gigaword corpus (which can be obtained from the Linguistic Dataset Consortium) together with the rest of our code.

\subsubsection{Downstream Training}

We train all models with $lr = 0.00005$. For each target ratio $r$ and each of $\transformer$ and $\transformerplusplus$, we select the best $\lambda_{len} \in \{0.1, 0.2, 0.5, 1, 2, 5, 10\}$ in terms of ROUGE-L on the validation set and report test set results in Table~\ref{tab:res-sentsum}.

\section{Additional Results}

\subsection{Yelp-Reviews}

\begin{figure*}
    \centering    
    \includegraphics[width=\columnwidth]{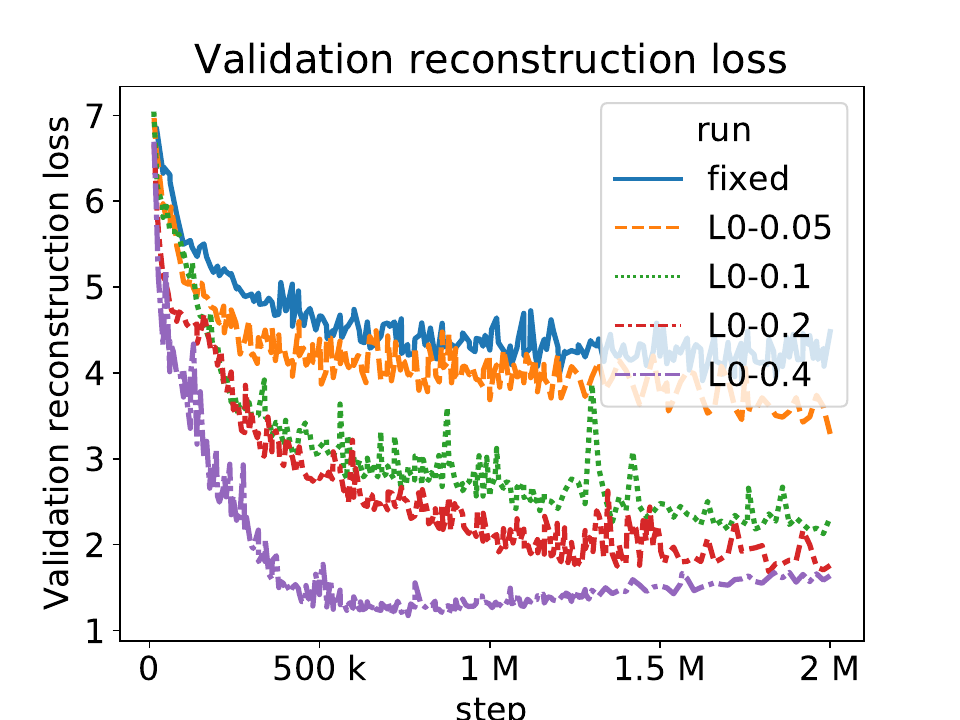}
    \caption{Reconstruction loss on the validation set of Yelp-Reviews for different autoencoders. \textbf{fixed}: The bag consists of a single vector obtained by averaging the embeddings at the last layer of the Transformer encoder. \textbf{L0-r}: BoV-AE with L0Drop target ratio $r$.}
    \label{fig:yelp-reviews-valrec}
\end{figure*}

In Figure~\ref{fig:yelp-reviews-valrec}, we plot the reconstruction ability of the fixed-size model compared to the BoV-AEs on the validation set.

Again, despite a large dimensionality ($d = 512$), the single-vector model achieves substantially lower reconstruction ability than BoV-AE. With respect to the target sparsity rate, we find that $r = 0.1$ is enough to reach dramatically better results than the fixed-size model, whereas $r = 0.05$ only reaches slightly better results after two million training steps. 
However, the plot shows clearly that \textbf{L0-0.05} has not converged, suggesting that \textbf{L0-0.05} could reach much better performance if trained for even longer.

\subsection{Yelp-Sentences}

\input{sections/window-size.tex}
\subsubsection{Computation time}\label{app:yelp-sentences-time}
Our experiments have shown that bag-of-vector representations are more powerful than single-vector representations. However, the increased capacity of BoV-AE comes at the expense of higher computation time. The size of the latent representation impacts the computation time in two places: During cross-attention in the decoder and when computing the mapping. Asymptotically, the decoder's cross-attention mechanism computes $\bigO(n \cdot |s|)$ dot-products, where $n$ is the number of vectors in the latent representation and $|s|$ is the length of the text sequence $s$. When computing the mapping, both at training and inference time, we produce a fixed number $N$ of vectors autoregressively, but in most applications, $N$ can reasonably be bound by a linear function of $n$ (e.g., $2n$ in style transfer or $0.5n$ in summarization). The mapping is essentially a Transformer decoder, so both the cross attention and self attention parts compute $\bigO(n^2)$ dot-products. Given that $n = 1$ for single-vector AEs and $n = \bigO(|s|)$ for BoV-AEs with L0Drop, we obtain the asymptotic complexities as shown in Table~\ref{tab:asymp-comp}.
\begin{table*}[]
    \centering
    \caption{Asymptotic computation time in the $\embtoemb$ framework as a function of the latent representation size $n$ and the length of the input text $|s|$, depending on the type of autoencoder.}
    \label{tab:asymp-comp}
    \begin{tabular}{c|c|c}
        AE type & Cross-Attention Decoding & Mapping \\
        \hline
        in general & $\bigO(n \cdot |s|)$ & $\bigO(n^2)$ \\
        fixed & $\bigO(|s|)$ & $\bigO(1)$ \\
        BoV-AE & $\bigO(|s|^2)$ & $\bigO(|s|^2)$ \\
    \end{tabular}
\end{table*}

To assess the empirical impact, we measure the wallclock time of $\embtoemb$'s "Inference" stage (cf. Figure~\ref{fig:emb2emb-framework}). We take separate measurements for encoding, mapping, and decoding, respectively. Since decoding speed depends on the quality of generation (e.g., when the end-of-sequence symbol is generated late due to repetitions), we do the following to enable fairer comparisons. We enforce the same fixed number of decoding steps (10) in all models. The mapping is set to produce as many output vectors as input vectors. We use a batch size of 1, but note that the results would largely extend to larger batch sizes when binned batching is used. The results are shown in Table~\ref{tab:model-speeds}.

Both the encoding and the mapping stages of $\embtoemb$ are more expensive in BoV models than in the fixed-size model. The difference in the encoding stage can be explained by the overhead through the L0Drop layer, which includes identifying near-zero gates and discarding their respective vectors. The difference in the mapping grows with higher L0Drop target ratios. This is expected since the number of autoregressive steps decreases with the target ratio.
Finally, we do not observe any meaningful speed differences between the models at decoding time. This is somewhat surprising, but could be explained by two factors. First, the \emph{self-attention} part of the decoder already has a complexity of $\bigO(|s|^2)$, which probably dominates the total computation time. Secondly, the computation of the dot-product is easy to parallelize. In summary, we find that BoV models are slower overall, especially in the mapping. However, since our L0Drop implementation prunes near-zero vectors, lower target rates mitigated the additional computation overhead. This is especially evident when comparing training speeds. While \textbf{L0-0.8} processes 15 sentences per second, \textbf{L0-0.4} processes can process 21 (fixed-size: 42).

\begin{table*}[]
    \centering
        \caption{The number of seconds it takes to process 5\% of the validation set (1264 samples) with a batch size of 1. Lower is better.}
    \label{tab:model-speeds}
    \begin{tabular}{c|c|c|c}
    Model & Encoding & Mapping & Decoding \\
    \hline
    fixed & 4.8 & 2.4 & 51.7 \\
    L0-0.4 & 7.2 & 12.6 & 50.1  \\
    L0-0.8 & 7.3 & 20.6 & 50.3 \\ 
    \end{tabular}
\end{table*}

\subsubsection{Using Pretrained Autoencoders}\label{app:yelp-sentences-pretraining}
The $\embtoemb$ framework is in principle compatible with any autoencoder.
This enables us to leverage large-scale pretraining, which has proven to be a very powerful method in NLP recently, e.g. with BERT~\citep{devlin-etal-2019-bert}. 
Due to the extremely high computational cost, training a large BoV-AE on a large general-purpose corpus is out of scope for this paper.
However, given the plug and play nature of $\embtoemb$, we can build on top of BART~\cite{lewis2019bart}, which uses similar resources as BERT, but is trained via a denoising autoencoder objective.
We can use this model either as is, or add an L0Drop layer between the encoder and decoder and finetune the model on our target dataset Yelp-Sentences.

For finetuning, we use the same training scheme as for our models, namely a denoising objective where we delete 10\% of the input tokens from the input at random. The model is trained through Adam with a learning rate of 0.00005. We use an L0Drop target rate of 0.4.
Our experimental results show that, when no L0Drop is used, the BART-based model gets to a validation reconstruction loss of 0.05 after only 5k training steps. This is a strong improvement over our best BoV models trained from scratch, which plateau at a loss of 1.0, demonstrating the power of large scale pretraining. With L0Drop, the model converges at roughly 0.29 after only 100k of finetuning, despite a relatively low target rate of 0.4.

When training on sentiment transfer downstream, we find the same pattern as for the models trained from scratch. If we don't finetune BART at all or finetune without L0Drop, downstream training is unable to learn to both retain a high self-BLEU score and achieve high transfer accuracy.
Whenever the transfer accuracy goes above 50\%, self-BLEU goes to very small scores ($< 1$). However, when L0Drop is used, the model achieves 35 points in self-BLEU at a target accuracy of 61\%. This confirms again our hypothesis that L0Drop regularization is needed to make the model work. In quantitative terms, BART with L0Drop is comparable to the BoV model \textbf{L0-0.4}, which was trained from scratch and achieves 55\% accuracy and 38 points self-BLEU.
Qualitatively, however, we observe that the pretrained model generates more fluent text. In Table~\ref{tab:yelp-sentences-bart-examples}, we show 10 randomly sampled examples of the model trained from scratch versus BART finetuned with L0Drop and a target rate of 0.4.
While both models are relatively good at retaining words from the input text, the pretrained model generally produces text that is more grammatical and coherent than the model trained from scratch (see examples \#1, \#2, \#3, \#6, \#9, \#10). This can be attributed to the language model of BART, which was pretrained to generate human-written text from a large general-purpose corpus.
Yet, the model outputs could clearly be improved further. We hypothesize that finetuning on a very domain-specific target dataset like Yelp-Sentences leads the model to quickly forget knowledge learned during pretraining, a phenomenon often observed with pretrained language models~\citep{yogatama2019learning}.
In the future, we would like to train a large BoV-AE model with L0Drop on a large general-purpose corpus, so that it can be used out of the box in the $\embtoemb$ framework for any task.

\begin{table*}[]
    \centering
        \caption{10 randomly sampled examples from Yelp-Sentences, evaluated on a BoV model trained with an L0Drop target rate of 0.4 from scratch versus a model initialized with BART and finetuned with L0Drop of 0.4.}
    \begin{tabularx}{\textwidth}{l|X||X|X}
        \# & \textbf{Input sentence} & \textbf{Output of L0-0.4} & \textbf{Output of BART with L0Drop} \\
        \hline
        \hline
1 & the restroom situation alone is enough for any woman to go crazy . & the restroom situation alone is enough for the woman to always good ! & great restroom and that alone is worth it.\\
\hline
2 & she would push my moms hands out of the way and just plain rude ! & she would gain out my hands out of the way and so wonderful ! & wow, they keep the ladies hands out!\\
\hline
3 & i hate it when it takes \_num\_ minutes to get a cup of coffee . & i makes maggie pointing it she mr. r ( , and wonderful ! & love it when it takes \_num\_ minutes to get.\\
\hline
4 & see update below . & see an frustrating . & see update below. \\
\hline
5 & the way they submitted the loan was false which caused the decline on purpose . & the receptionist they always the inspection and she caused the stage is always ! & the way they made the sale was very. \\
\hline
6 & another bad italian take out story . & another bad italian of take new notch . & great, good italian pizza.\\
\hline
7 & if you want a refrigerator , that 'll be  \_num\_ extra . & if for ajo sons picky ' ' ' mien hemmed and huge ! & great place, you 'll get a great.\\
\hline
8 & get new staff , they were just terrible ! & get the new staff , they were always terrible ! & great food, great staff!\\
\hline
9 & i recently visited while searching for a venue for a commitment ceremony and reception . & i found brake while select for venue for a workout and and wonderful ! & wow, i recently visited this location for a wedding.\\
\hline
10 & this place is why yelp should allow zero stars . & this place is that yelp who should not great ! & this place is great if you love starbucks.\\

    \end{tabularx}
    \label{tab:yelp-sentences-bart-examples}
\end{table*}

